\documentclass{article}

\usepackage[preprint]{neurips_2026}

\usepackage[utf8]{inputenc} 
\usepackage[T1]{fontenc}    
\usepackage{hyperref}       
\usepackage{url}            
\usepackage{booktabs}       
\usepackage{amsfonts}       
\usepackage{nicefrac}       
\usepackage{microtype}      
\usepackage{booktabs}
\usepackage{subcaption}    
\usepackage[dvipsnames]{xcolor}   
\usepackage{xcolor}         
\usepackage{colortbl}
\usepackage{graphicx}
\usepackage{multirow}
\usepackage{xcolor}
\usepackage{adjustbox}
\usepackage{enumitem}
\usepackage{amsmath}      
\usepackage{graphicx}
\definecolor{dartmouthgreen}{RGB}{0,105,62}

\title{Beyond MoCap: Scaling Motion Tokenizers with Synthetic Human Motion for Generative Modeling}

\author{
Yiwen Yan \qquad
Wanning He \qquad
Yu-Wing Tai \\
Dartmouth College
}

\begin{document}

\maketitle

\begin{abstract}
Human motion generation models are fundamentally constrained by the limited diversity of motion capture datasets, which predominantly contain common, repetitive actions and fail to cover the long tail of complex human movements, resulting in a restricted motion vocabulary in learned latent representations and poor generalization to rare, compositional, and highly dynamic motions. In this work, we propose a framework for expanding the motion representation space by leveraging large-scale synthetic human motion, introducing a data generation pipeline that produces diverse, physically plausible motion sequences beyond the distribution of existing datasets and integrating it with a redesigned VQ-VAE tokenizer that adapts to this expanded motion space. Unlike conventional tokenizers trained on narrow data distributions, our approach jointly scales both the training distribution and the discrete codebook, enabling the model to capture a significantly richer set of motion primitives. We demonstrate that training with synthetic motion substantially improves the coverage and compositionality of the learned motion vocabulary, leading to consistent gains across motion generation tasks such as text-to-motion and motion continuation, while remaining fully compatible with existing frameworks including MotionGPT. Our results suggest that the primary bottleneck lies in the limited support of the learned motion representation, rather than model architecture alone. Scaling synthetic motion in tandem with representation learning offers a principled path toward more expressive, controllable, and generalizable human motion synthesis.
\end{abstract}
\section{Introduction}

\begin{figure}[t]
  \centering
  \includegraphics[width=\linewidth]{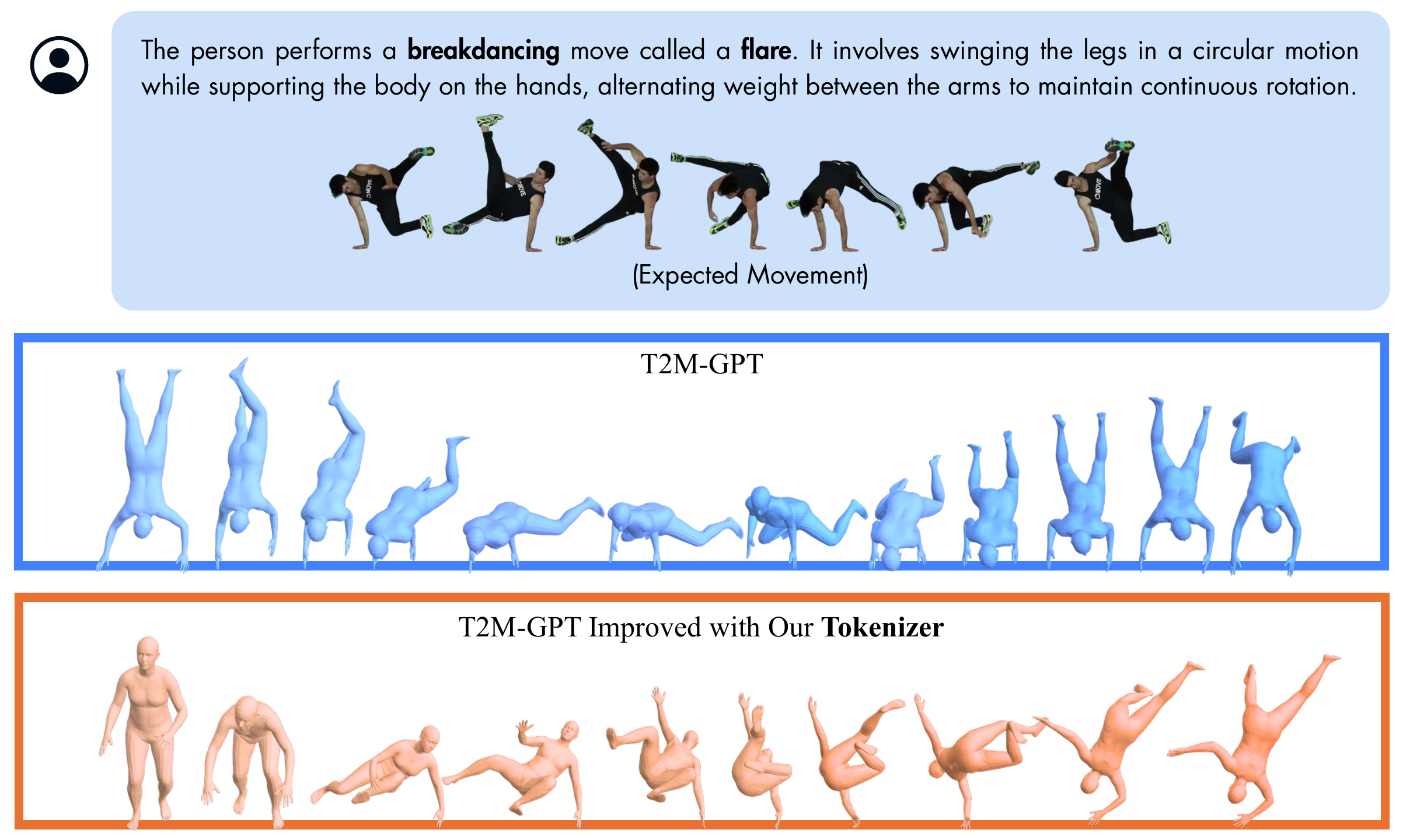}\\
  \vspace{-0.1in}
  \caption{Given a textual description of a complex human motion, existing text-to-motion models such as T2M-GPT~\cite{zhang2023t2mgpt} fail to faithfully reproduce the intended dynamics due to the limited expressiveness of MoCap-trained tokenizers. By contrast, our approach leverages large-scale synthetic motion and an enhanced VQ-VAE tokenizer to expand the motion vocabulary, enabling the model to synthesize motions that closely match the expected motions. This highlights that the primary bottleneck lies in the restricted token space, and that scaling motion representations is key to improving generative performance.}
  \vspace{-0.2 in}
  \label{fig:overview}
\end{figure}

Recent advances in human motion generation, particularly those built upon discrete latent representations such as VQ-VAE tokenizers~\cite{oord2017vqvae} and autoregressive transformers (e.g., MotionGPT~\cite{zhang2024motiongpt}), have demonstrated strong performance in tasks including text-to-motion~\cite{guo2022generating,zhang2024motiongpt} and motion continuation. However, despite rapid architectural progress, these systems remain fundamentally limited by the quality and diversity of their learned motion tokenizers. In existing pipelines, the tokenizer is trained on motion capture (MoCap) datasets such as Human3.6M~\cite{ionescu2013human3} and AMASS~\cite{mahmood2019amass}, which are inherently narrow in distribution, dominated by everyday actions such as walking, sitting, and simple interactions. As a result, the learned codebooks capture only a restricted subset of human motion, leading to poor representation of rare, complex, or highly dynamic movements. This limitation propagates to downstream generative models, which are consequently unable to synthesize motions that lie outside the support of the tokenizer, regardless of model capacity.

A key challenge is the difficulty of scaling motion capture data. Collecting high-quality MoCap is expensive, labor-intensive, and requires specialized equipment and controlled environments, making it impractical to cover the full spectrum of human motion, especially rare, dangerous, or highly creative actions such as acrobatics or stylized performances. As a result, motion datasets exhibit limited diversity and poor long-tail coverage~\cite{guo2022generating}. Simply scaling model size or training time cannot overcome this data bottleneck, since the learned motion vocabulary remains constrained by the training distribution.

This limitation highlights the central role of the motion tokenizer in modern generative pipelines. As the interface between continuous motion signals and discrete models, the tokenizer defines the effective motion vocabulary available for generation. A limited tokenizer restricts expressiveness, whereas a richer and more structured token space improves diversity, compositionality, and controllability. Similar trends have been observed in vision and language, where stronger discrete representations directly enhance generative modeling~\cite{esser2021taming,razavi2019vqvae2}. Therefore, advancing motion generation requires not only better architectures but also more expressive and scalable motion representations (Fig.~\ref{fig:overview}).

In this work, we propose to address this challenge by jointly expanding the motion data distribution and the tokenizer capacity through large-scale synthetic human motion. We introduce a scalable motion generation pipeline that synthesizes diverse and physically plausible motion sequences beyond the scope of existing MoCap datasets. To ensure physical plausibility, we incorporate constraints derived from kinematics, dynamics, and contact consistency, following common practices in motion synthesis and physics-aware modeling~\cite{holden2016deep,rempe2021humor}. At the same time, we promote diversity by systematically exploring the long tail of motion space, generating rare, extreme, and highly compositional sequences that are underrepresented or absent in real-world datasets. Building upon this enriched data distribution, we redesign the VQ-VAE tokenizer with an expanded and better-utilized codebook, allowing it to encode a significantly broader spectrum of motion primitives.


To transfer the improved motion representation to downstream generation, we adapt existing autoregressive motion generation frameworks to the new motion token space while keeping their original architectures unchanged.
In these models, motion is represented as a sequence of discrete tokens and generated autoregressively through next-token prediction, in close analogy to language modeling. We therefore re-tokenize the motion side of the training data with our augmented tokenizer and continue fine-tuning pretrained generators on the resulting token sequences. This design serves two purposes: it enables a direct assessment of how improved tokenization affects downstream generation, and it provides a simple practical recipe for upgrading existing discrete-token motion generation models with a better tokenizer.

Extensive experiments demonstrate that our method substantially improves motion diversity, realism, and compositionality across multiple benchmarks. In text-to-motion generation, our approach produces more complex and expressive motions that better match detailed textual descriptions. In motion continuation tasks, it enables more coherent and dynamic long-term predictions. We further show that our tokenizer achieves higher codebook utilization and better coverage of motion space, validating the effectiveness of scaling both data and representation.

\noindent\textbf{Contributions.}
We summarize our contributions as follows:
\begin{itemize}[leftmargin=20pt, itemsep=2pt, topsep=2pt, parsep=0pt, partopsep=0pt]
\item We identify the limited support of motion tokenizers as a key bottleneck in modern motion generation, where restricted motion vocabularies induced by narrow MoCap data constrain diversity and compositionality.

\item We highlight the importance of jointly scaling the motion data distribution and token capacity, showing that improving generation requires expanding both rather than scaling model architectures alone.

\item We introduce a scalable synthetic motion generation pipeline that expands the long tail of motion space while preserving physical plausibility.

\item We develop an augmented VQ-VAE tokenizer that adapts to the expanded distribution, achieving better codebook utilization and a richer motion vocabulary.

\item We demonstrate consistent improvements across multiple generation models without architectural changes, improving both in-distribution fidelity and out-of-distribution generalization.

\end{itemize}
\vspace{-0.05in}
\section{Related Work}
\vspace{-0.05in}

\noindent\textbf{Human Motion Generation Models.} Recent advances in human motion generation have improved realism and controllability. However, most methods assume the training distribution adequately covers the target motion space, which fails for rare, complex, or weakly represented actions.

Early language-to-motion works, including Text2Action~\cite{ahn2017text2action}, Language2Pose~\cite{ahuja2019language2pose}, and Action2Motion~\cite{guo2020action2motion}, established text-motion mappings but were limited by small datasets and simple models. Modern datasets such as KIT Motion-Language~\cite{plappert2016kit} and the CMU Motion Capture Database provide richer supervision, yet still suffer from long-tail limitations. Existing approaches include diffusion-based models~\cite{tevet2022motionclip,shafir2023priormdm,chen2023mld} and discrete-token autoregressive frameworks~\cite{zhang2023t2mgpt,guo2024momask}, achieving strong results on benchmarks such as HumanML3D~\cite{guo2022generating}. Yet, scaling model capacity alone yields diminishing returns when the data distribution is limited~\cite{lu2025scamo}, and fails when target motions lie outside its support. To improve semantic alignment, recent works integrate large language models, such as MotionGPT~\cite{jiang2023motiongpt} and Motion-Agent~\cite{wu2025motionagent}. While effective at high-level consistency, they do not address motion representation and remain constrained by available motion primitives. Compositional approaches~\cite{athanasiou2022teach,shafir2023priormdm} improve long-horizon structure but are similarly bounded by motion coverage.

In contrast to prior work focusing on architecture or conditioning, we identify the \emph{motion tokenizer} as a key bottleneck. Limited data induces a restricted motion vocabulary, constraining the motions that generative models can represent. This motivates jointly expanding the training distribution and the representation space.

\noindent\textbf{Discrete Motion Representations and Tokenization.}
Discrete motion representations enable language-model-style generation by mapping continuous motion into token sequences. Most methods adopt VQ-VAE-based tokenizers~\cite{oord2017vqvae}, as used in MotionGPT~\cite{jiang2023motiongpt} and T2M-GPT~\cite{zhang2023t2mgpt}, typically with small codebooks (e.g., $\sim$512 entries)~\cite{zhang2023t2mgpt,jiang2023motiongpt,guo2024momask}. However, human motion exhibits richer temporal structure and stronger long-tail variability, making such limited codebooks a representational bottleneck. Diverse motions are compressed into a small set of discrete codes, reducing expressiveness and leading to under-utilization or collapse, often mitigated by heuristics such as EMA updates or code resetting~\cite{zhang2023t2mgpt}. Recent work identifies token capacity as a key scaling factor. ScaMo~\cite{lu2025scamo} shows performance depends jointly on model, data, and token scale, while MotionCtrl~\cite{cao2025motionctrl}, MotionRL~\cite{liu2024motionrl}, and MotionChain~\cite{jiang2024motionchain} highlight the role of richer representations in improving controllability and reasoning.

Building on this idea, we look at motion tokenization from the perspective of scaling data. When we expand the motion distribution with synthetic data, standard codebooks are no longer enough. To address this, we increase the size of the codebook, study how well it works through ablation experiments, and introduce a loss-weighting strategy to keep it stable and avoid collapse. This allows the model to better cover the larger motion space and produce improved results.

\noindent\textbf{Data Limitations and Synthetic Motion Generation} Data augmentation is well studied in pose-related tasks, especially in static or frame-wise settings. Methods such as PoseAug~\cite{gong2021poseaug} and EvoSkeleton~\cite{li2020cascaded} expand pose distributions via spatial transformations or structural perturbations. Extending these techniques to motion sequences is non-trivial, as valid motion requires temporal coherence, physical plausibility, and long-range consistency. Sequence-level augmentation for motion generation remains underexplored. MotionAug~\cite{maeda2022motionaug} focuses on motion prediction rather than generation. Recent data-scaling and self-improving pipelines~\cite{gillman2024selfcorrecting,guo2025snapmogen,cao2025motionctrl} emphasize dataset quality and scale, but rely mainly on curation or filtering rather than generating new motion sequences.

In contrast, we introduce a synthetic motion generation pipeline that expands the range of motions the model can learn. It works on a structured representation that keeps motions physically valid while allowing controlled randomness, so it can generate diverse and realistic sequences, including rare and combined motions that do not appear in datasets such as AMASS~\cite{mahmood2019amass} and KIT~\cite{plappert2016kit}. When combined with a scaled tokenizer, this data increases the effective motion vocabulary for downstream models, allowing them to generate motions beyond the original training distribution and addressing a limitation not handled by prior work.
\section{Methods}
\label{sec:method}

\noindent\textbf{Overview.}
We address the limited motion vocabulary induced by MoCap data by jointly scaling
(1) the motion data distribution and (2) the capacity of the motion tokenizer.
Our framework consists of two key components:
(i) a synthetic motion generation pipeline that expands the coverage of human motion space,
and (ii) an augmented VQ-VAE tokenizer that adapts to this expanded distribution.
The resulting tokenizer can be directly integrated into existing motion generation models.

\noindent
Our design is guided by three principles:
(1) \textit{validity}: generated motions must remain physically plausible;
(2) \textit{diversity}: the pipeline should explore the long-tail of motion space;
(3) \textit{compatibility}: synthesized data should align with the representation used by downstream models.


\subsection{Motion Synthesis}

\begin{figure}[t]
  \centering
  \vspace{-0.1in}
  \includegraphics[width=\linewidth]{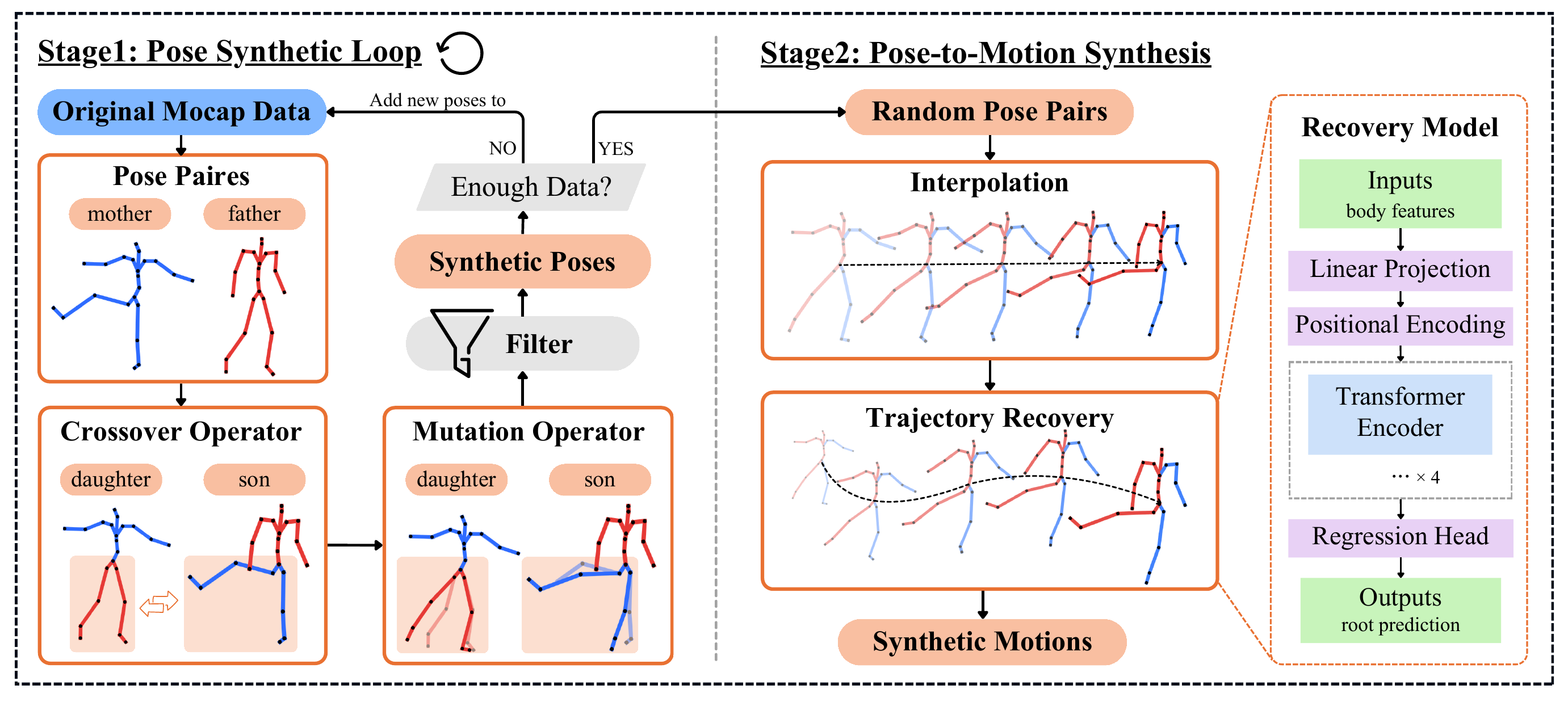}\\
  \vspace{-0.1in}
  \caption{Overview of the motion synthesis pipeline. We generate diverse poses via crossover and mutation, filter them using a physical prior, and construct motion sequences through interpolation and trajectory recovery.}
  \vspace{-0.2in}
  \label{fig:datapipeline}
\end{figure}

Our motion synthesis pipeline consists of two main stages (Fig.~\ref{fig:datapipeline}):
(1) pose synthesis via structured stochastic exploration, and
(2) pose-to-motion conversion for generating motion sequences.

The pose synthesis stage is inspired by genetic algorithms~\cite{holland1992adaptation} and builds upon~\cite{li2020cascaded}. Generated poses are filtered using a dynamics pose prior~\cite{akhter2015pose}, ensuring physically plausible skeletons.
We then construct motion sequences by interpolating between valid poses and recover global trajectories using a lightweight transformer.
The pipeline is implemented in the SMPL representation~\cite{loper2023smpl} and converted to HumanML3D format~\cite{guo2022generating}.

\noindent\textbf{Hierarchical Pose Representation.}
We represent each pose as a kinematic tree rooted at the pelvis, where each bone is parameterized 
in a local coordinate frame using spherical coordinates. This representation decouples orientation 
from bone length, enabling structured manipulation while preserving anatomical consistency.

\noindent\textbf{Structured Exploration via Crossover and Mutation.}
To expand the pose distribution, we draw inspiration from genetic algorithms and perform 
stochastic exploration through crossover and mutation operations. 
Given two poses, crossover exchanges subtrees at randomly selected limb roots (e.g., arms or legs), 
recombining semantically meaningful body parts while preserving kinematic structure. 
Mutation further perturbs bone orientations with Gaussian noise while keeping bone lengths fixed.

Importantly, these operations are not arbitrary perturbations: they preserve the compositional 
structure of human motion while enabling exploration of configurations that are unlikely to 
appear in MoCap data. This allows the pipeline to systematically populate underrepresented 
regions of the pose space.

\noindent\textbf{Pose Filtering.}
To ensure physical plausibility, we enforce anatomical constraints using a pose prior that restricts 
joint angles to feasible ranges. Invalid poses are discarded. This step ensures that the expanded 
distribution remains within the manifold of valid human poses, preventing the tokenizer from 
modeling unrealistic artifacts.

\noindent\textbf{Pose-to-Motion Conversion.} We construct motion sequences by interpolating between valid poses using spherical interpolation. 
While simple, this process plays a critical role: it exposes the model to transitions between 
poses that are rare or absent in real datasets, thereby expanding the coverage of motion dynamics. 
This effectively enriches the transition manifold, which is essential for generative models that 
compose motion tokens over time.

\noindent\textbf{Trajectory Recovery.}
The interpolated motions are represented in a pelvis-centered coordinate frame, where the root remains stationary in the global space. We recover their global trajectory based on the underlying motion dynamics, using the recovery model shown in Fig.~\ref{fig:datapipeline}.

We represent a motion sequence by decoupling the root information from the rest of the body as
\begin{equation}
\mathbf{m}_{1:T} = \{\mathbf{m}_t\}_{t=1}^T, \quad \mathbf{m}_t = [\mathbf{r}_t, \mathbf{b}_t] \in \mathbb{R}^{263},
\end{equation}
where $\mathbf{r}_t \in \mathbb{R}^{4}$ denotes the root features, consisting of frame-wise rotational and planar translational increments, along with the root height, i.e., $\mathbf{r}_t = (\omega_t, v^x_t, v^z_t, h_t)$; $\mathbf{b}_t \in \mathbb{R}^{259}$ represents the body features. 

We use the body component $\mathbf{b}_{1:T}$ as input to predict the corresponding root dynamics. The model first projects the input body features into a latent space, followed by positional encoding and a multi-layer Transformer encoder~\cite{vaswani2017attention} to capture long-range temporal dependencies. The final root features are obtained through a lightweight prediction head.

Given the predicted root sequence, the global trajectory is recovered in two components. The root orientation is obtained by integrating the angular velocity, while the global position is computed by accumulating the rotated planar velocities. 


\subsection{Motion Tokenization}
\label{sec:tokenizer}

We adopt a vector-quantized variational autoencoder (VQ-VAE)~\cite{guo2022generating, zhang2023t2mgpt} to discretize motion sequences into a finite set of motion tokens, enabling language-model-style generation.

Given a motion sequence $\mathbf{m}_{1:T} \in \mathbb{R}^{T \times D}$, an encoder $E$ maps it to a latent representation:
\begin{equation}
\mathbf{z}_{1:T/N} = E(\mathbf{m}_{1:T}),
\end{equation}
where $N$ is the temporal downsampling factor.

\noindent\textbf{Vector Quantization.}
Each latent vector $\mathbf{z}_t$ is quantized by nearest-neighbor lookup in a codebook $\mathcal{C} = \{\mathbf{c}_k\}_{k=1}^{K}$:
\begin{equation}
\hat{\mathbf{z}}_t = \arg\min_{\mathbf{c}_k \in \mathcal{C}} \| \mathbf{z}_t - \mathbf{c}_k \|_2^2.
\end{equation}

The quantized sequence $\hat{\mathbf{z}}_{1:T/N}$ is then decoded to reconstruct the motion:
\begin{equation}
\hat{\mathbf{m}}_{1:T} = D(\hat{\mathbf{z}}_{1:T/N}).
\end{equation}

\noindent\textbf{Training Objective.}
The tokenizer is trained using a combination of reconstruction and commitment losses:
\begin{equation}
\mathcal{L}_{\text{vq}} =
\| \mathbf{m} - \hat{\mathbf{m}} \|_1 
+ \alpha \| \mathbf{p} - \hat{\mathbf{p}} \|_1 
+ \beta \| \mathbf{z} - \mathrm{sg}[\hat{\mathbf{z}}] \|_2^2,
\end{equation}
where $\mathbf{p}$ denotes joint positions, $\mathrm{sg}[\cdot]$ is the stop-gradient operator, and $\alpha, \beta$ are weighting factors.
We adopt exponential moving average (EMA) updates and periodic codebook reset strategies following~\cite{zhang2023t2mgpt} to improve codebook utilization.

\paragraph{Augmented Training with Synthetic Motions.}
To expand the coverage of motion patterns, we train the tokenizer on a mixture of real and synthesized data:
\begin{equation}
\mathcal{D} = (1 - \lambda)\mathcal{D}_{\text{real}} + \lambda \mathcal{D}_{\text{syn}},
\end{equation}
where $\lambda$ controls the contribution of synthetic data.

This augmentation exposes the tokenizer to rare and compositional motion patterns that are underrepresented in MoCap datasets, improving its ability to represent the long-tail of motion space.

\paragraph{Codebook Scaling.}
As the diversity of the training distribution increases, a fixed-size codebook becomes insufficient to represent fine-grained motion primitives.
We therefore increase the codebook size $K$ to match the expanded distribution. From a quantization perspective, a larger codebook reduces reconstruction error by providing a finer partition of the latent space.
However, excessively large $K$ may lead to under-utilization.
In practice, we identify an operating regime where codebook utilization remains high while reconstruction quality and downstream performance improve.

By jointly scaling the training data and codebook capacity, the tokenizer transitions from a bottleneck to a flexible interface between motion signals and generative models.
This results in a richer and more compositional motion vocabulary, which directly benefits downstream motion generation.

\subsection{Adapting Motion Generation Models}
\noindent\textbf{Autoregress Motion Generation.}
Autoregressive motion generation models cast motion synthesis as discrete
sequence modeling. Given a motion sequence $\mathbf{x} = (x_1, \dots,
x_T)$ in continuous space, a motion tokenizer first maps it into a
sequence of discrete codes $\mathbf{z} = (z_1, \dots, z_L)$, where each
$z_\ell \in \{1, \dots, K\}$ indexes an entry in a learned codebook of
size $K$. The generator then models the token sequence autoregressively as
\begin{equation}
p(\mathbf{z}\mid \mathbf{c}) = \prod_{\ell=1}^{L} p(z_\ell \mid z_{<\ell},
\mathbf{c}),
\end{equation}
where $\mathbf{c}$ denotes the conditioning signal, such as a text
description. Motion generation is thus reduced to next-token prediction in
a discrete vocabulary, analogous to autoregressive language modeling.


\noindent\textbf{Training Procedure.}
We first re-tokenize the training motions with our augmented tokenizer and use the resulting token sequences as the new supervision targets, while keeping the paired text unchanged. Starting from the pretrained checkpoint, we then continue fine-tuning the original autoregressive generator without changing its architecture. Notably, the amount of training data remains the same, as we only reuse existing text-motion pairs with the new tokenizer for fine-tuning. The newly synthesized motions are not used in this training, since we do not generate new text pairs for them.



\section{Experiments}
\label{sec:experiments}
We design our experiments to answer three questions.
\textbf{(i)} Does our synthetic data expand the motion distribution
beyond MoCap, and do the resulting motions remain anatomically valid?
\textbf{(ii)} Does training on this expanded distribution yield a
better tokenizer, both in- and out-of-distribution?
\textbf{(iii)} Does the improved tokenizer transfer to downstream
generation models without changing their architecture?
We address each question in
Sec.~\ref{sec:exp_data},
Sec.~\ref{sec:exp_tokenizer},
and Sec.~\ref{sec:exp_sota} respectively.

\subsection{Experiment Setup}
\label{sec:exp_setup}

\noindent\textbf{HumanML3D.} 
HumanML3D~\cite{guo2022generating} is a large-scale 3D human motion dataset, which contains 14,616 motion sequences and paired text annotations. 
We build our synthetic motion data based on the entire HumanML3D  set and generate approximately $64\times$ additional synthetic motions using our synthesis pipeline, resulting in a large-scale augmented dataset. We use HumanML3D as the baseline to analyze the distributional differences introduced by our synthesized data.
We train the motion tokenizer on the augmented HumanML3D dataset. For downstream motion generation, we adopt the standard $80\%/5\%/15\%$ train/val/test split and train models on the training set.

\noindent\textbf{Motion-X.} 
Motion-X++~\cite{zhang2025motion} is a large-scale multimodal 3D whole-body human motion dataset, which contains 120.5K motion sequences spanning diverse scenarios such as music, kungfu, and performance. 
Compared to HumanML3D, Motion-X++ exhibits significantly higher diversity in motion patterns and semantic annotations, making it a challenging benchmark for evaluating generalization. We use Motion-X++ solely for evaluation without further training to assess the cross-dataset generalization capability of our tokenizer and downstream generation models.

\noindent\textbf{Evaluation metrics.}
For tokenizer evaluation, we report \emph{Mean Per Joint Position Error (MPJPE)} for reconstruction accuracy.
For motion generation evaluation, following~\cite{guo2022generating}, we use the pre-trained motion/text feature extractor of~\cite{guo2022generating} and report five standard metrics:
\emph{R-Precision} (Top-1, Top-2, Top-3) for text-motion retrieval consistency;
\emph{Fréchet Inception Distance (FID)} for distributional fidelity;
\emph{Multimodal Distance (MM-Dist)} for text-motion feature alignment;
and \emph{Diversity} for generation variety.
For each metric, we repeat the evaluation $20$ times and report the average with $95\%$ confidence interval.

\subsection{Synthetic Data Evaluation}
\label{sec:exp_data}
\begin{figure}[t]
  \centering
  \includegraphics[width=\linewidth]{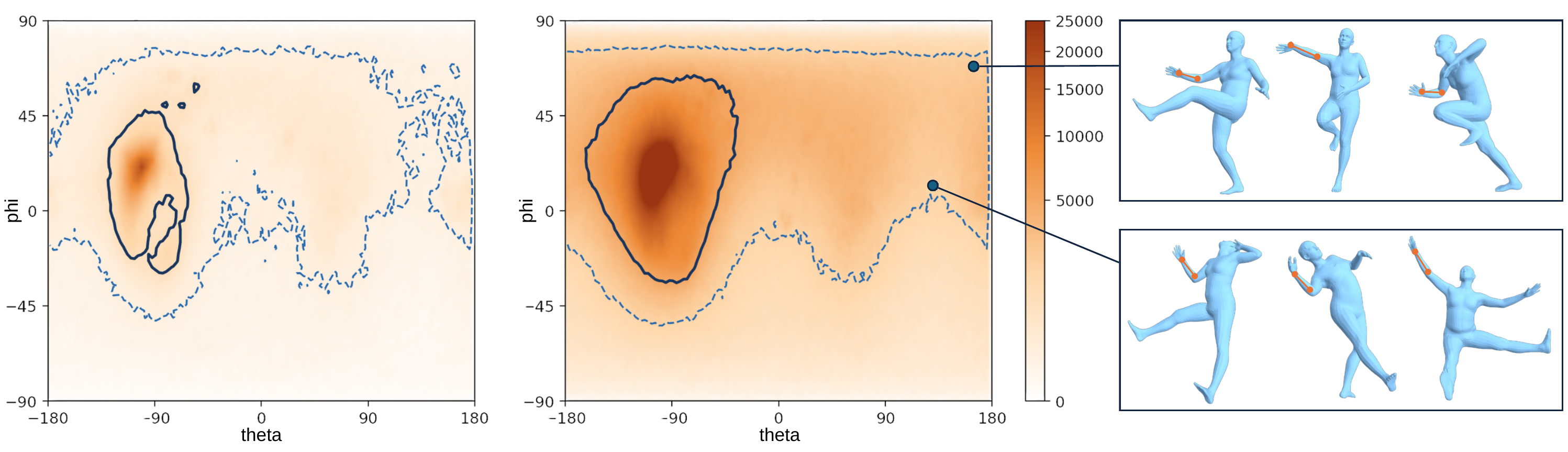}\\
  \vspace{-0.05in}
  \caption{
  \textbf{Bone Orientation Distribution (Elbow to Wrist).}
  We visualize the angular distribution of the right forearm  using spherical coordinates, where the horizontal axis represents $\theta$ and the vertical axis represents $\phi$. 
  \textbf{Left:} HumanML3D. \textbf{Right:} Our synthesized data. 
  The solid contours encloses the core 50\% of the data distribution, while the dashed contour encloses 90\% of the samples, representing the majority of the dataset.
  }
  \vspace{-0.15in}
  \label{fig:distribution}
\end{figure}

We assess our synthetic data along two axes: 
\textbf{coverage} of the motion distribution and 
\textbf{validity} of the resulting poses. These two measurements establish the precondition for the 
rest of our experiments

\noindent\textbf{Coverage.}
We examine the angular distribution of every bone vector in the SMPL skeleton under each bone's local coordinate frame. Aggregated over all 21 bones, every single bone shows a non-trivial expansion, and none of them shrinks.
Fig.~\ref{fig:distribution} visualizes one representative bone, showing that the synthetic distribution covers a substantially larger region of orientations and populates areas that are sparsely visited or entirely absent in HumanML3D.
Detailed overage statistics are provided in App.~\ref{app:bone_coverage}, Tab.~\ref{tab:bone_stats}

\textbf{Validity.}
Coverage is only useful if the new poses remain anatomically valid.
We filter synthesized poses using the dynamics pose prior~\cite{akhter2015pose} and retain those whose validity score is no lower than that of their parent pose. Under this criterion, 67.33\% of generated poses are kept for training.
We sample poses on the right of Fig.~\ref{fig:distribution} from the newly explored regions, which indicate that the expanded coverage corresponds to plausible, though relatively rare, human configurations rather than anatomically impossible ones.

\subsection{Evalution of Motion Tokenizer}
\label{sec:exp_tokenizer}

\begin{table}[h]
\centering
\small
\begin{subtable}{0.45\linewidth}
\centering
\vspace{-0.1in}
\label{tab:tok-id}
\begin{tabular}{lcc}
\toprule
Metric & T2M-GPT& \textbf{Ours}\\
\midrule
FID $\downarrow$        & 0.132 & \textbf{0.076}\,{\scriptsize\textcolor{ForestGreen}{($-42.4\%$)}} \\
Top-1 $\uparrow$        & 0.499 & \textbf{0.504}\,{\scriptsize\textcolor{red}{($+1.0\%$)}}  \\
MM-Dist $\downarrow$    & 3.011 & \textbf{2.975}\,{\scriptsize\textcolor{ForestGreen}{($-1.2\%$)}}  \\
Diversity $\uparrow$    & 9.764 & \textbf{9.804}\,{\scriptsize\textcolor{red}{($+0.4\%$)}}  \\
\bottomrule
\end{tabular}
\caption{In-distribution on HumanML3D}
\end{subtable}
\hfill
\begin{subtable}{0.52\linewidth}
\centering
\vspace{-0.1in}
\label{tab:tok-ood}
\begin{tabular}{lccc}
\toprule
Subset & \# samples & T2M-GPT $\downarrow$ & \textbf{Ours} $\downarrow$ \\
\midrule
Overall & 25{,}865 & 272.73 & \textbf{252.37}\,{\scriptsize\textcolor{ForestGreen}{($-7.5\%$)}}  \\
Daily   & 12{,}042 & 244.70 & \textbf{231.98}\,{\scriptsize\textcolor{ForestGreen}{($-5.2\%$)}}  \\
Sports  &  6{,}944 & 322.65 & \textbf{282.74}\,{\scriptsize\textcolor{ForestGreen}{($-12.4\%$)}} \\
Dance   &  3{,}394 & 180.77 & \textbf{161.80}\,{\scriptsize\textcolor{ForestGreen}{($-10.5\%$)}} \\
\bottomrule
\end{tabular}
\caption{Out-of-
distribution on Motion-X++ subsets}
\end{subtable}
\vspace{-0.05in}
\caption{Our tokenizer enhances T2M-GPT both in-distribution on HumanML3D, and out-of-distribution on Motion-X++ subsets unseen during training. The complete per-subset breakdown in (b) is reported in App.~\ref{app:per_subset}.}
\vspace{-0.15in}
\label{tab:tokenizer}
\end{table}



\textbf{In-distribution.}
On HumanML3D, reconstruction FID drops from $0.132$ to $\mathbf{0.076}$  ($\mathbf{-42.4}\%$) while every retrieval metric improves (Tab.~\ref{tab:tokenizer}\,(a)).

\textbf{Out-of-distribution.}
We evaluate the tokenizer on the unseen dataset Motion-X++ and its task-specific subsets, which lie farther from the support of training data. 
As shown in Tab.~\ref{tab:tokenizer}\,(b), MPJPE drops on all tasks, suggesting that the expanded data improves the tokenizer's generalization and representation capacity by extending coverage beyond the original HumanML3D support, rather than merely densifying the existing distribution.



\subsection{Evaluation of Motion Generation}
\label{sec:exp_sota}

\begin{table*}[h]
\centering
\setlength{\tabcolsep}{4pt}
\begin{adjustbox}{max width=\textwidth, center}
\begin{tabular}{l c ccc c c c}
\toprule
\multirow{2}{*}{Method}
  & FID $\downarrow$
  & \multicolumn{3}{c}{R-Precision $\uparrow$}
  & MM-Dist $\downarrow$
  & Diversity $\uparrow$
  & MModality $\uparrow$ \\
\cmidrule(lr){3-5}
 & & Top-1 & Top-2 & Top-3 & & & \\
\midrule
MDM~\cite{tevet2022mdm}
  & $0.544^{\pm.044}$
  & $0.320^{\pm.005}$ & $0.498^{\pm.004}$ & $0.611^{\pm.007}$
  & $5.566^{\pm.027}$
  & $9.559^{\pm.086}$
  & $2.799^{\pm.072}$ \\
MLD~\cite{chen2023mld}
  & $0.473^{\pm.013}$
  & $0.481^{\pm.003}$ & $0.673^{\pm.003}$ & $0.772^{\pm.002}$
  & $3.196^{\pm.010}$
  & $9.724^{\pm.082}$
  & $2.413^{\pm.079}$ \\
MotionDiffuse~\cite{zhang2024motiondiffuse}
  & $0.630^{\pm.001}$
  & $0.491^{\pm.001}$ & $0.681^{\pm.001}$ & $0.782^{\pm.001}$
  & $3.113^{\pm.001}$
  & $9.410^{\pm.049}$
  & $1.553^{\pm.042}$ \\
T2M~\cite{guo2022generating}
  & $1.067^{\pm.002}$
  & $0.457^{\pm.002}$ & $0.559^{\pm.007}$ & $0.740^{\pm.003}$
  & $3.340^{\pm.008}$
  & $9.188^{\pm.002}$
  & $2.090^{\pm.083}$ \\
TM2T~\cite{guo2022tm2t}
  & $1.501^{\pm.017}$
  & $0.424^{\pm.003}$ & $0.618^{\pm.003}$ & $0.729^{\pm.002}$
  & $3.467^{\pm.011}$
  & $8.589^{\pm.076}$
  & $2.424^{\pm.093}$ \\
MoMask~\cite{guo2024momask}
  & $0.045^{\pm.002}$
  & $0.521^{\pm.002}$ & $0.713^{\pm.002}$ & $0.807^{\pm.002}$
  & $2.958^{\pm.008}$
  & $9.620^{\pm.064}$
  & $1.241^{\pm.040}$ \\
MotionChain~\cite{jiang2024motionchain}
  & $0.248^{\pm.009}$
  & $0.504^{\pm.003}$ & $0.617^{\pm.002}$ & $0.790^{\pm.003}$
  & $3.033^{\pm.010}$
  & $9.470^{\pm.075}$
  & $1.727^{\pm.014}$ \\
MotionGPT-2~\cite{wang2024motiongpt}
  & $0.191^{\pm.004}$
  & $0.496^{\pm.002}$ & $0.691^{\pm.003}$ & $0.782^{\pm.004}$
  & $3.080^{\pm.013}$
  & $9.860^{\pm.026}$
  & $2.137^{\pm.022}$ \\
Motion-R1~\cite{ouyang2025motion}
  & $0.201^{\pm.004}$
  & $0.515^{\pm.003}$ & $0.719^{\pm.002}$ & $0.818^{\pm.002}$
  & $2.854^{\pm.010}$
  & $10.206^{\pm.075}$
  & $2.137^{\pm.105}$ \\

\midrule
T2M-GPT~\cite{zhang2023t2mgpt}
  & $0.116^{\pm.004}$
  & ${\mathbf{0.491^{\pm.003}}}$ 
  & ${\mathbf{0.680^{\pm.003}}}$ 
  & ${\mathbf{0.775^{\pm.002}}}$
  & ${\mathbf{3.118^{\pm.011}}}$ 
  & ${\mathbf{9.761^{\pm.081}}}$ 
  & $1.856^{\pm.011}$ \\
\rowcolor{green!15}
T2M-GPT (Ours)
  & $\mathbf{0.097^{\pm.005}}$   
  & $0.464^{\pm.003}$            
  & $0.647^{\pm.002}$   
  & $0.747^{\pm.002}$            
  & $3.271^{\pm.009}$            
  & $9.594^{\pm.081}$            
  & $\mathbf{{2.266}^{\pm.095}}$ \\
\midrule
MotionGPT~\cite{jiang2023motiongpt}
  & $0.232^{\pm.008}$
  & $\mathbf{{0.492^{\pm.003}}}$ 
  & $\mathbf{{0.681^{\pm.003}}}$ 
  & $\mathbf{{0.778^{\pm.002}}}$
  & $\mathbf{{3.096^{\pm.008}}}$ 
  & $9.528^{\pm.071}$ 
  & $2.008^{\pm.084}$ \\
\rowcolor{green!15}
MotionGPT (Ours)
  & $\mathbf{0.176^{\pm.007}}$
  & $0.432^{\pm.003}$
  & $0.611^{\pm.003}$           %
  & $0.710^{\pm.003}$
  & $3.578^{\pm.011}$
  & $\mathbf{9.578^{\pm.082}}$
  & $\mathbf{{4.845}^{\pm.163}}$ \\
\midrule
MotionAgent~\cite{wu2025motionagent}
  & $0.230^{\pm.009}$
  & $\mathbf{{0.515}^{\pm.004}}$ 
  & $\mathbf{{0.691}^{\pm.003}}$ 
  & $\mathbf{{0.801}^{\pm.004}}$
  & $\mathbf{{2.967}^{\pm.020}}$ 
  & $\mathbf{{9.908}^{\pm.102}}$ 
  & $2.142^{\pm.014}$ \\
\rowcolor{green!15}
MotionAgent (Ours)
  & $\mathbf{0.184^{\pm.009}}$   
  & $0.453^{\pm.003}$            
  & $0.637^{\pm.002}$            
  & $0.738^{\pm.002}$            
  & $3.318^{\pm.009}$            
  & $9.665^{\pm.076}$            
  & $\mathbf{2.904^{\pm.230}}$ \\
\bottomrule
\end{tabular}
\end{adjustbox}
\vspace{-0.1in}
\caption{\textbf{Comparison on HumanML3D.}
The green rows indicate results with our synthetic data and augmented tokenizer; bold values indicate the better result within each baseline/ours pair.}
\vspace{-0.15in}
\label{tab:humanml3d}
\end{table*}
\begin{figure}[h]
  \centering
  \includegraphics[width=\linewidth]{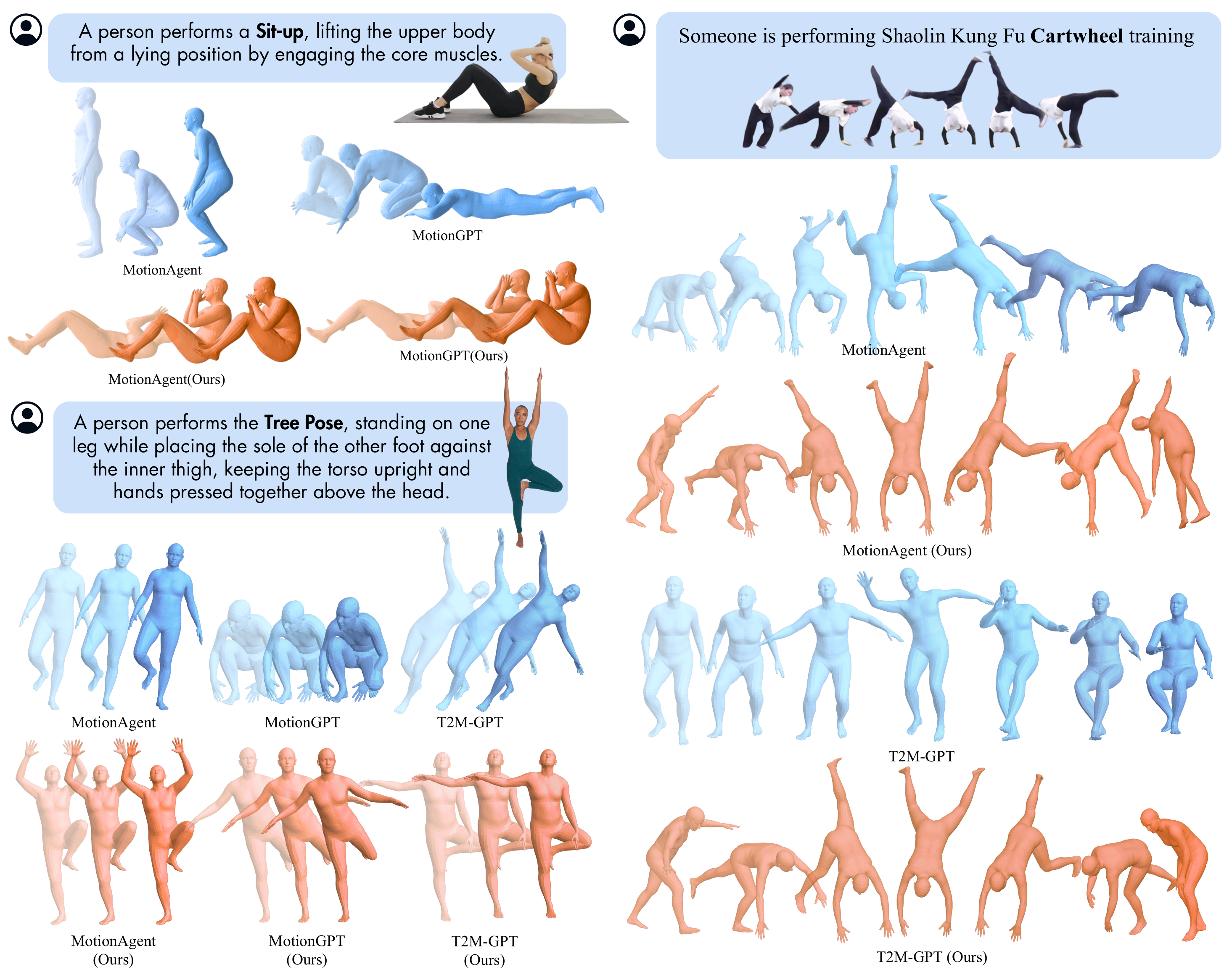}\\
  \vspace{-0.1in}
  \caption{
  Qualitative motion generation results on challenging prompts, including rare actions such as sports, yoga, and kung-fu. For each example, the original model output is shown in blue, and the result from our method is shown in orange below.
  }
  \vspace{-0.15in}
  \label{fig:generation}
\end{figure}


\noindent\textbf{Quantitative results.} Tab.~\ref{tab:humanml3d} shows a consistent pattern: our synthetic-data training and augmented tokenizer mainly improve motion fidelity. After replacing the original tokenizer while keeping the generator architecture unchanged, FID drops from $0.116$ to $0.097$ for T2M-GPT ($\mathbf{16.4\%}$), from $0.232$ to $0.176$ for MotionGPT ($\mathbf{24.1\%}$), and from $0.230$ to $0.184$ for MotionLLM ($\mathbf{20.0\%}$). This improvement is consistent across all three backbones, suggesting that the gain comes from better motion representation rather than model-specific tuning. We note a slight drop in text–motion alignment. This is likely due to bias in the HumanML3D test set, as illustrated in Fig.~\ref{fig:distribution}. Overall, our method enhances motion realism and diversity while keeping other metrics comparable, supporting tokenizer scaling for better generation.


\paragraph{Qualitative Results.}

We further present qualitative visualization results in Fig.~\ref{fig:generation}. The
example prompts include uncommon motions such as sports, yoga, and kung-fu. These are out-of-distribution prompts that do not exist in the training data. In each
example, the result improved with our method is shown in orange, below the original model
result shown in blue.   
Our method produces visibly better motions, supporting our main claim that the proposed motion synthesis pipeline broadens motion-space coverage and improves the generalization of downstream generation models. More qualitative examples can be found in the App.~\ref{app:more_qualitative}

\newcommand{\best}[1]{\textcolor{red}{#1}}
\newcommand{\second}[1]{\textcolor{blue}{#1}}

\section{Conclusion}

In this work, we present a motion data synthesis pipeline, Beyond MoCap, that both expands motion-space coverage and scales the codebook accordingly.
This simple change improves tokenizer reconstruction quality on both in-distribution and out-of-distribution 
generalization tasks, and transfers consistently to downstream motion generation models. 
These results demonstrate that better motion generation is not solely a problem of scaling the generator, but also requires scaling the motion distribution and the tokenizer together.

As with any other work, our method has several limitations. Our synthetic pipeline currently focuses only on body motion, without modeling finer-grained hand or facial dynamics. Furthermore, we focus on expanding motion diversity 
without pairing the synthesized motions with corresponding text annotations. While current expansion significantly improves the realism and diversity metrics of motion generation, it is foreseeable that incorporating text labels would further enhance the semantic alignment capability of the model, leading to more comprehensive improvements across all 
aspects of generation performance.

\bibliographystyle{plainnat}
\bibliography{references}

@inproceedings{ahn2017text2action,
  title={Text2Action: Generative Adversarial Synthesis from Language to Action},
  author={Ahn, Hyemin and Ha, Timothy and Choi, Yunho and Yoo, Hwiyeon and Oh, Songhwai},
  booktitle={IEEE/CVF International Conference on Computer Vision (ICCV)},
  year={2017}
}

@inproceedings{ahuja2019language2pose,
  title={Language2Pose: Natural Language Grounded Pose Forecasting},
  author={Ahuja, Chaitanya and Morency, Louis-Philippe},
  booktitle={International Conference on 3D Vision (3DV)},
  year={2019}
}

@inproceedings{akhter2015pose,
  title={Pose-conditioned joint angle limits for 3D human pose reconstruction},
  author={Akhter, Ijaz and Black, Michael J},
  booktitle = {IEEE/CVF Conference on Computer Vision and Pattern Recognition (CVPR)},
  year={2015}
}

@inproceedings{athanasiou2022teach, 
  title = {{TEACH}: {T}emporal {A}ction {C}ompositions for {3D} {H}umans},
  author = {Athanasiou, Nikos and Petrovich, Mathis and Black, Michael J. and Varol, G{\"u}l},
  booktitle = {{International Conference on 3D Vision (3DV)}},
  year = {2022} 
  }

@inproceedings{cao2025motionctrl,
  author    = {Cao, Bin and Zheng, Sipeng and Wang, Ye and Xia, Lujie and Wei, Qianshan and Jin, Qin and Liu, Jing and Lu, Zongqing},
  title     = {MotionCtrl: A Real-time Controllable Vision-Language-Motion Model},
  booktitle = {IEEE International Conference on Computer Vision (ICCV)},
  year      = {2025},
}

@inproceedings{chen2023mld,
  author    = {Chen, Xin and Jiang, Biao and Liu, Wen and Huang, Zilong and Fu, Bin and Chen, Tao and Yu, Gang},
  title     = {Executing Your Commands via Motion Diffusion in Latent Space},
  booktitle = {IEEE/CVF Conference on Computer Vision and Pattern Recognition (CVPR)},
  year      = {2023},
}

@inproceedings{esser2021taming,
  title={Taming Transformers for High-Resolution Image Synthesis},
  author={Esser, Patrick and Rombach, Robin and Ommer, Bjorn},
  booktitle = {IEEE/CVF Conference on Computer Vision and Pattern Recognition (CVPR)},
  year={2021}
}

@inproceedings{gillman2024selfcorrecting,
  author    = {Gillman, Nate and Freeman, Michael and Aggarwal, Daksh and Hsu, Chia-Hong and Luo, Calvin and Tian, Yonglong and Sun, Chen},
  title     = {Self-Correcting Self-Consuming Loops for Generative Model Training},
  booktitle = {Proceedings of the International Conference on Machine Learning (ICML)},
  year      = {2024},
}

@inproceedings{gong2021poseaug,
  title       = {PoseAug: A Differentiable Pose Augmentation Framework for 3D Human Pose Estimation},
  author      = {Gong, Kehong and Zhang, Jianfeng and Feng, Jiashi},
  booktitle = {IEEE/CVF Conference on Computer Vision and Pattern Recognition (CVPR)},
  year        = {2021}
}

@inproceedings{guo2020action2motion,
  title={Action2Motion: Conditioned Generation of 3D Human Motions},
  author={Guo, Chuan and Zuo, Xinxin and Wang, Sen and Zou, Shihao and Sun, Qingyao and Deng, Annan and Gong, Minglun and Cheng, Li},
  booktitle={ACM International Conference on Multimedia (MM)},
  year={2020}
}

@inproceedings{guo2022generating,
  title={Generating diverse and natural 3d human motions from text},
  author={Guo, Chuan and Zou, Shihao and Zuo, Xinxin and Wang, Sen and Ji, Wei and Li, Xingyu and Cheng, Li},
  booktitle={IEEE/CVF Conference on Computer Vision and Pattern Recognition (CVPR)},
  year={2022}
}

@inproceedings{guo2024momask,
  author    = {Guo, Chuan and Mu, Yuxuan and Javed, Muhammad Gohar and Wang, Sen and Cheng, Li},
  title     = {MoMask: Generative Masked Modeling of 3D Human Motions},
  booktitle = {IEEE/CVF Conference on Computer Vision and Pattern Recognition (CVPR)},
  year      = {2024},
}

@inproceedings{guo2022tm2t,
  title={Tm2t: Stochastic and tokenized modeling for the reciprocal generation of 3d human motions and texts},
  author={Guo, Chuan and Zuo, Xinxin and Wang, Sen and Cheng, Li},
  booktitle={European Conference on Computer Vision (ECCV)},
  year={2022},
}

@inproceedings{guo2025snapmogen,
  author    = {Guo, Chuan and Hwang, Inwoo and Wang, Jian and Zhou, Bing},
  title     = {SnapMoGen: Human Motion Generation from Expressive Texts},
 booktitle={Advances in Neural Information Processing Systems (NeurIPS)},
  year      = {2025},
}

@inproceedings{holden2016deep,
  title={A deep learning framework for character motion synthesis and editing},
  author={Daniel Holden and Jun Saito and Taku Komura},
  booktitle={SIGGRAPH},
  year={2016}
}

@book{holland1992adaptation,
  title={Adaptation in natural and artificial systems: an introductory analysis with applications to biology, control, and artificial intelligence},
  author={Holland, John H},
  publisher={MIT press},
  year={1992}
}

@article{ionescu2013human3,
  title={Human3. 6m: Large scale datasets and predictive methods for 3d human sensing in natural environments},
  author={Ionescu, Catalin and Papava, Dragos and Olaru, Vlad and Sminchisescu, Cristian},
  journal={IEEE TPAMI},
  year={2013},
}

@inproceedings{jiang2023motiongpt,
  title={Motiongpt: Human motion as a foreign language},
  author={Jiang, Biao and Chen, Xin and Liu, Wen and Yu, Jingyi and Yu, Gang and Chen, Tao},
  booktitle={Advances in Neural Information Processing Systems (NeurIPS)},
  year={2024}
}

@inproceedings{jiang2024motionchain,
  title={Motionchain: Conversational motion controllers via multimodal prompts},
  author={Jiang, Biao and Chen, Xin and Zhang, Chi and Yin, Fukun and Li, Zhuoyuan and Yu, Gang and Fan, Jiayuan},
  booktitle={European Conference on Computer Vision (ECCV)},
  year={2024},
}

@inproceedings{li2020cascaded,
  title={Cascaded deep monocular 3d human pose estimation with evolutionary training data},
  author={Li, Shichao and Ke, Lei and Pratama, Kevin and Tai, Yu-Wing and Tang, Chi-Keung and Cheng, Kwang-Ting},
  booktitle = {IEEE/CVF Conference on Computer Vision and Pattern Recognition (CVPR)},
  year={2020}
}

@article{liu2024motionrl,
  title={MotionRL: Align Text-to-Motion Generation to Human Preferences with Multi-Reward Reinforcement Learning}, 
  author={Xiaoyang Liu and Yunyao Mao and Wengang Zhou and Houqiang Li},
  year={2024},
  journal={arXiv preprint arXiv:2410.06513},
}

@incollection{loper2023smpl,
  title={{SMPL}: A skinned multi-person linear model},
  author={Loper, Matthew and Mahmood, Naureen and Romero, Javier and Pons-Moll, Gerard and Black, Michael J},
  booktitle={Seminal Graphics Papers: Pushing the Boundaries, Volume 2},
  year={2023}
}

@inproceedings{lu2025scamo,
  author    = {Lu, Shunlin and Wang, Jingbo and Lu, Zeyu and Chen, Ling-Hao and Dai, Wenxun and Dong, Junting and Dou, Zhiyang and Dai, Bo and Zhang, Ruimao},
  title     = {ScaMo: Exploring the Scaling Law in Autoregressive Motion Generation Model},
  booktitle = {IEEE/CVF Conference on Computer Vision and Pattern Recognition (CVPR)},
  year      = {2025},
}

@inproceedings{maeda2022motionaug,
  author    = {Maeda, Takahiro and Ukita, Norimichi},
  title     = {MotionAug: Augmentation With Physical Correction for Human Motion Prediction},
  booktitle = {IEEE/CVF Conference on Computer Vision and Pattern Recognition (CVPR)},
  year      = {2022},
}

@inproceedings{mahmood2019amass,
  title={AMASS: Archive of Motion Capture as Surface Shapes}, 
  author={Naureen Mahmood and Nima Ghorbani and Nikolaus F. Troje and Gerard Pons-Moll and Michael J. Black},
  booktitle = {IEEE/CVF International Conference on Computer Vision (ICCV)},
  year={2019}
}

@inproceedings{oord2017vqvae,
  title={Neural Discrete Representation Learning},
  author={van den Oord, Aaron and Vinyals, Oriol and Kavukcuoglu, Koray},
 booktitle={Advances in Neural Information Processing Systems (NeurIPS)},
  year={2017}
}

@article{ouyang2025motion,
  title={Motion-R1: Enhancing Motion Generation with Decomposed Chain-of-Thought and RL Binding},
  author={Ouyang, Runqi and Li, Haoyun and Zhang, Zhenyuan and Wang, Xiaofeng and Zhang, Zeyu and Zhu, Zheng and Huang, Guan and Han, Sirui and Wang, Xingang},
  journal={International Conference on Learning Representations (ICLR)},
  year={2026}
}

@misc{plappert2016kit,
  title={The KIT Motion-Language Dataset},
  author={Plappert, Matthias and Mandery, Christian and Asfour, Tamim},
  note={Big Data Journal, 4(4):236--252},
  year={2016},
}

@inproceedings{razavi2019vqvae2,
  title={Generating Diverse High-Fidelity Images with VQ-VAE-2},
  author={Razavi, Ali and van den Oord, Aaron and Vinyals, Oriol},
  booktitle={Advances in Neural Information Processing Systems (NeurIPS)},
  year={2019}
}

@inproceedings{rempe2021humor,
  title={HuMoR: 3D Human Motion Model for Robust Pose Estimation}, 
      author={Davis Rempe and Tolga Birdal and Aaron Hertzmann and Jimei Yang and Srinath Sridhar and Leonidas J. Guibas},
  booktitle = {IEEE/CVF International Conference on Computer Vision (ICCV)},
  year={2021}
}

@inproceedings{shafir2023priormdm,
  title={Human Motion Diffusion as a Generative Prior},
  author={Shafir, Yoni and Tevet, Guy and Kapon, Roy and Bermano, Amit Haim},
  booktitle={International Conference on Learning Representations (ICLR)},
  year={2024}
}

@article{tevet2022mdm,
  author       = {Tevet, Guy and Raab, Sigal and Gordon, Brian and Shafir, Yonatan and Cohen-Or, Daniel and Bermano, Amit H.},
  title        = {Human Motion Diffusion Model},
  journal      = {arXiv preprint arXiv:2209.14916},
  year         = {2022}
}

@inproceedings{tevet2022motionclip,
  title={Motionclip: Exposing human motion generation to clip space},
  author={Tevet, Guy and Gordon, Brian and Hertz, Amir and Bermano, Amit H and Cohen-Or, Daniel},
  booktitle={European Conference on Computer Vision (ECCV)},
  year={2022}
}

@inproceedings{vaswani2017attention,
  title={Attention is all you need},
  author={Vaswani, Ashish and Shazeer, Noam and Parmar, Niki and Uszkoreit, Jakob and Jones, Llion and Gomez, Aidan N and Kaiser, {\L}ukasz and Polosukhin, Illia},
  booktitle={Advances in Neural Information Processing Systems (NeurIPS)},
  year={2017}
}

@article{wang2024motiongpt,
  title={Motiongpt-2: A general-purpose motion-language model for motion generation and understanding},
  author={Wang, Yuan and Huang, Di and Zhang, Yaqi and Ouyang, Wanli and Jiao, Jile and Feng, Xuetao and Zhou, Yan and Wan, Pengfei and Tang, Shixiang and Xu, Dan},
  journal={arXiv preprint arXiv:2410.21747},
  year={2024}
}

@inproceedings{wu2025motionagent,
  author    = {Wu, Qi and Zhao, Yubo and Wang, Yifan and Liu, Xinhang and Tai, Yu-Wing and Tang, Chi-Keung},
  title     = {Motion-Agent: A Conversational Framework for Human Motion Generation with LLMs},
  booktitle={International Conference on Learning Representations (ICLR)},
  year      = {2025},
}

@article{zhang2024motiondiffuse,
  title={Motiondiffuse: Text-driven human motion generation with diffusion model},
  author={Zhang, Mingyuan and Cai, Zhongang and Pan, Liang and Hong, Fangzhou and Guo, Xinying and Yang, Lei and Liu, Ziwei},
  journal={IEEE TPAMI},
  year={2024},
}

@inproceedings{zhang2023t2mgpt,
  author    = {Zhang, Jianrong and Zhang, Yangsong and Cun, Xiaodong and Zhang, Yong and Zhao, Hongwei and Lu, Hongtao and Shen, Xi and Shan, Ying},
  title     = {Generating Human Motion From Textual Descriptions With Discrete Representations},
  booktitle = {IEEE/CVF Conference on Computer Vision and Pattern Recognition (CVPR)},
  year      = {2023},
}

@inproceedings{zhang2024motiongpt,
  title={Motiongpt: Finetuned llms are general-purpose motion generators},
  author={Zhang, Yaqi and Huang, Di and Liu, Bin and Tang, Shixiang and Lu, Yan and Chen, Lu and Bai, Lei and Chu, Qi and Yu, Nenghai and Ouyang, Wanli},
  booktitle={Proceedings of the AAAI Conference on Artificial Intelligence},
  year={2024}
}

@article{zhang2025motion,
  title={Motion-x++: A large-scale multimodal 3d whole-body human motion dataset},
  author={Zhang, Yuhong and Lin, Jing and Zeng, Ailing and Wu, Guanlin and Lu, Shunlin and Fu, Yurong and Cai, Yuanhao and Zhang, Ruimao and Wang, Haoqian and Zhang, Lei},
  journal={arXiv preprint arXiv:2501.05098},
  year={2025}
}

\newpage
\appendix
\section{Overview} 
In this appendix, we present:
\begin{itemize}
\item Section~\ref{app:more_qualitative}: more qualitative results.
\item Section~\ref{app:bone_coverage}: per-bone coverage statistics that quantify how the synthetic data expands beyond the original HumanML3D support.
\item Section~\ref{app:codebook_ablation}: ablation study of the
codebook size.
\item Section~\ref{app:ratio_ablation}: ablation study of the mixing
ratio between original and synthetic data.
\item Section~\ref{app:impl}: implementation details of the data
generation, tokenizer training and finetuning.
\item Section~\ref{app:eval}: additional details on the evaluation
metrics and the motion representation.
\item Section~\ref{app:per_subset}: full per-subset out-of-distribution
reconstruction results on Motion-X++.
\end{itemize}

\section{More Qualitative Results}
\label{app:more_qualitative}
\begin{center}
\includegraphics[width=\linewidth]{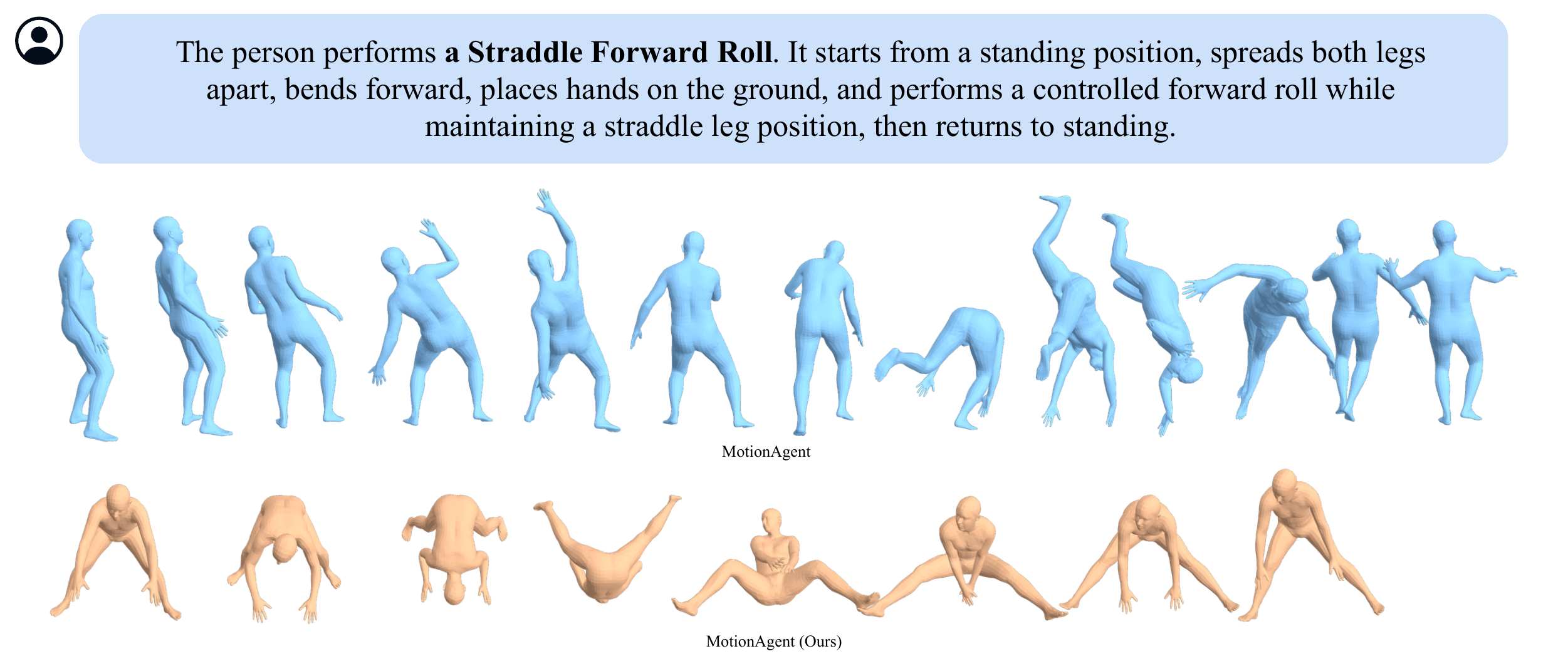}
\includegraphics[width=\linewidth]{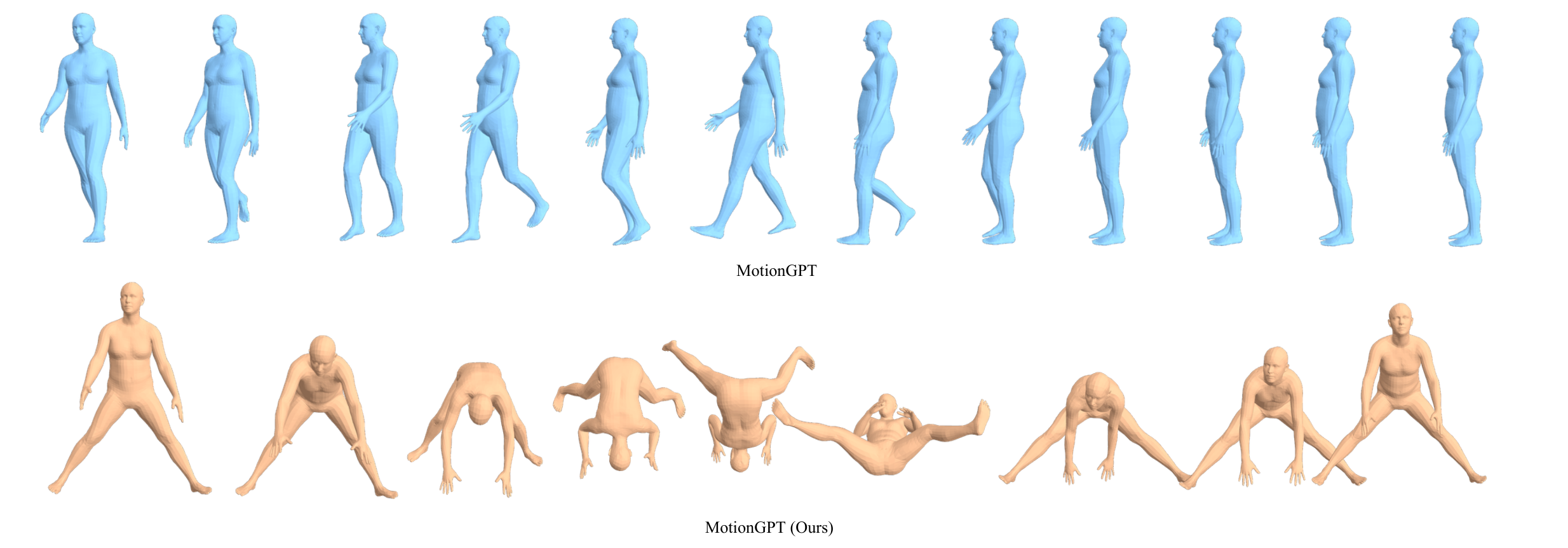}
\includegraphics[width=\linewidth]{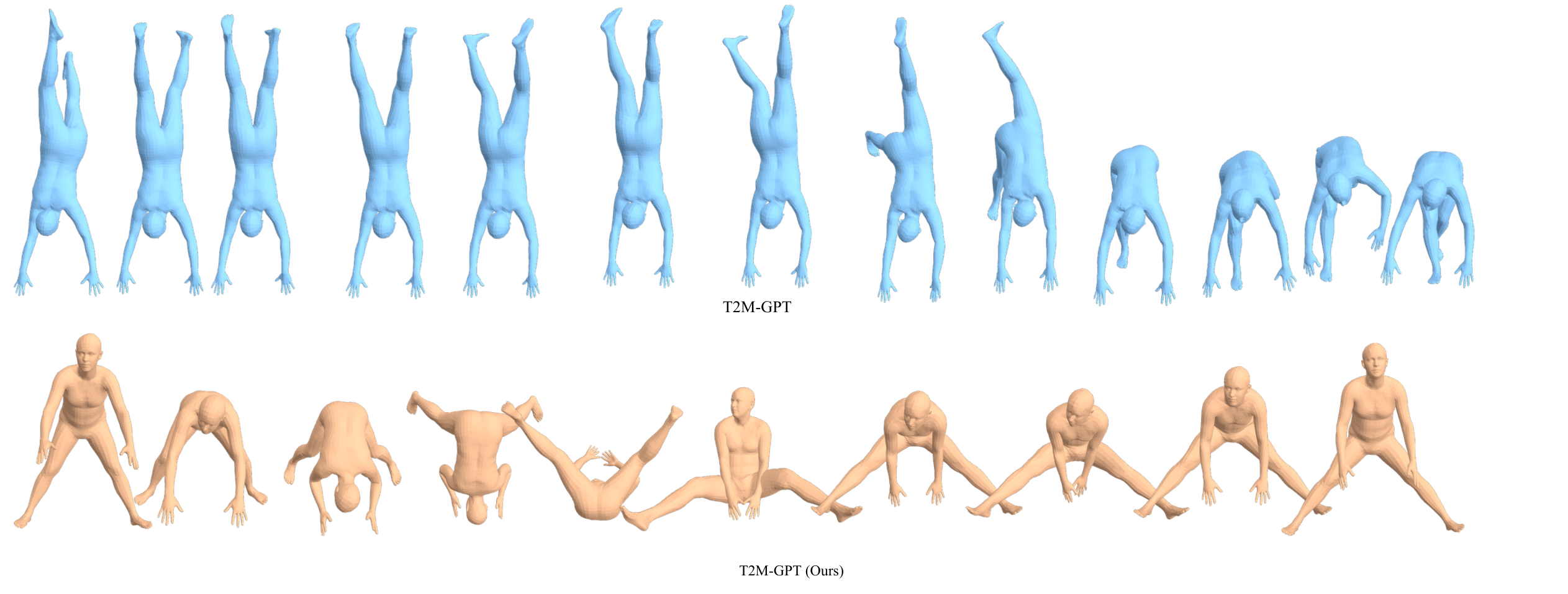}
\includegraphics[width=\linewidth]{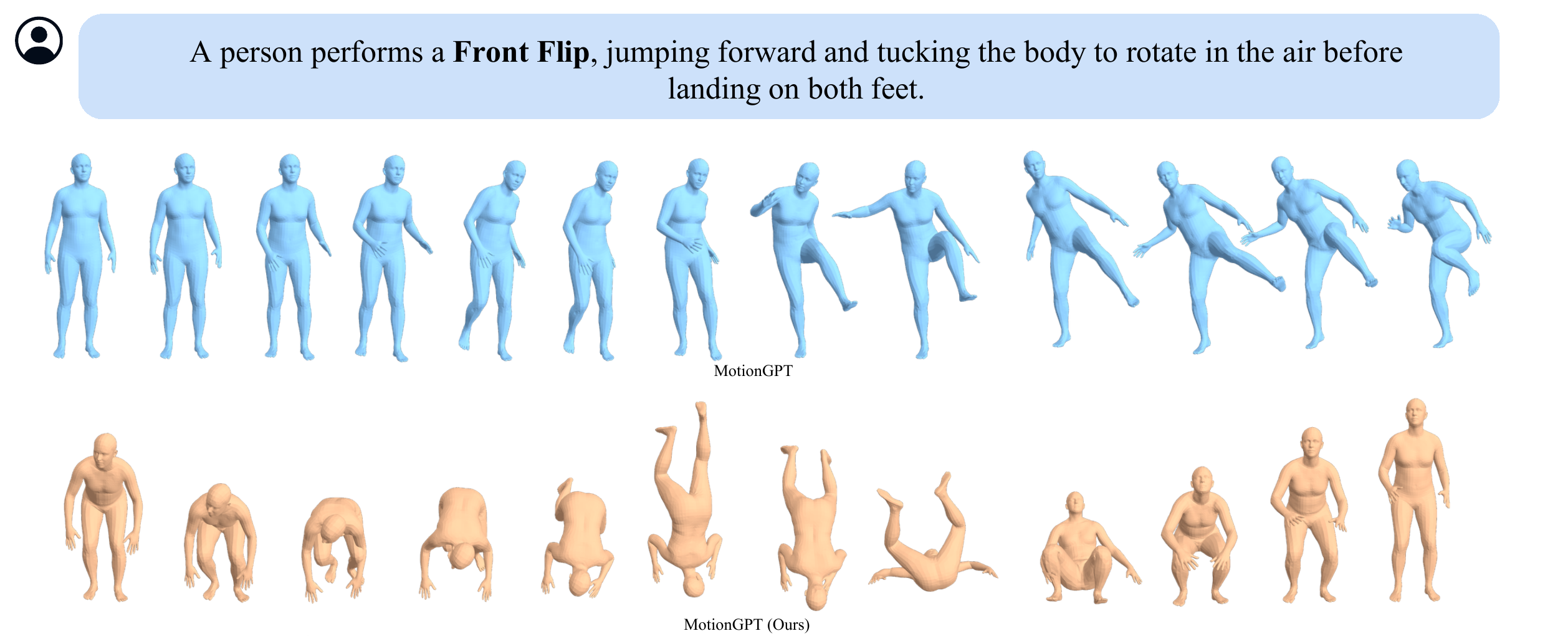}
\includegraphics[width=\linewidth]{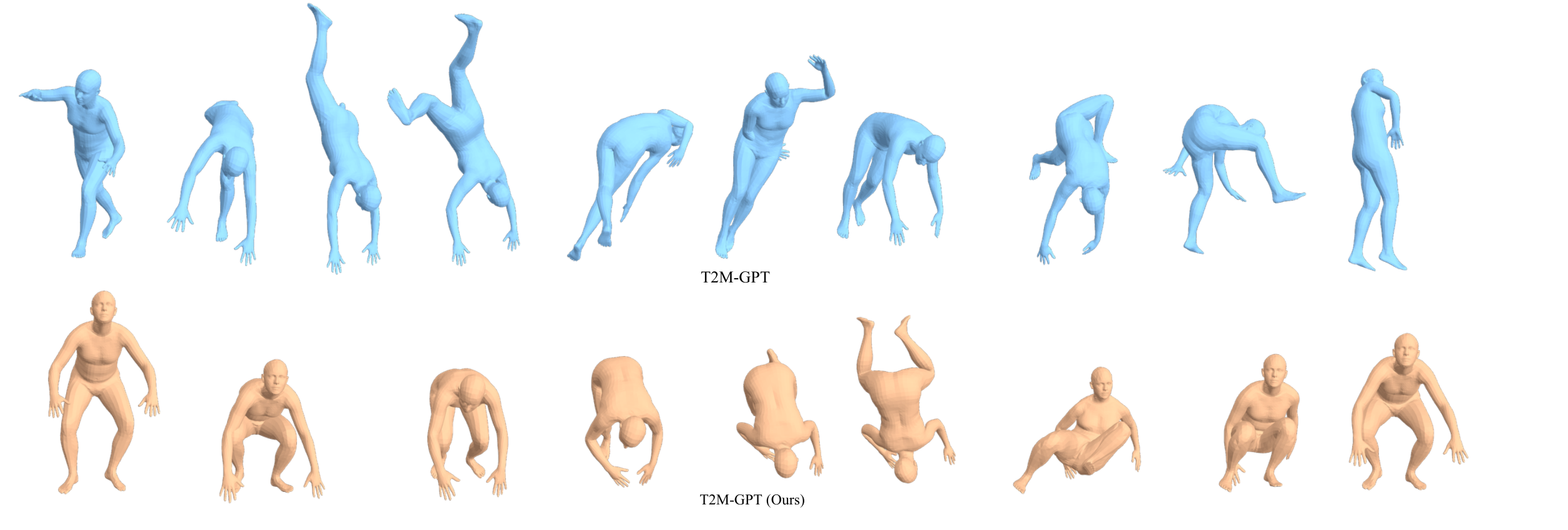}
\includegraphics[width=\linewidth]{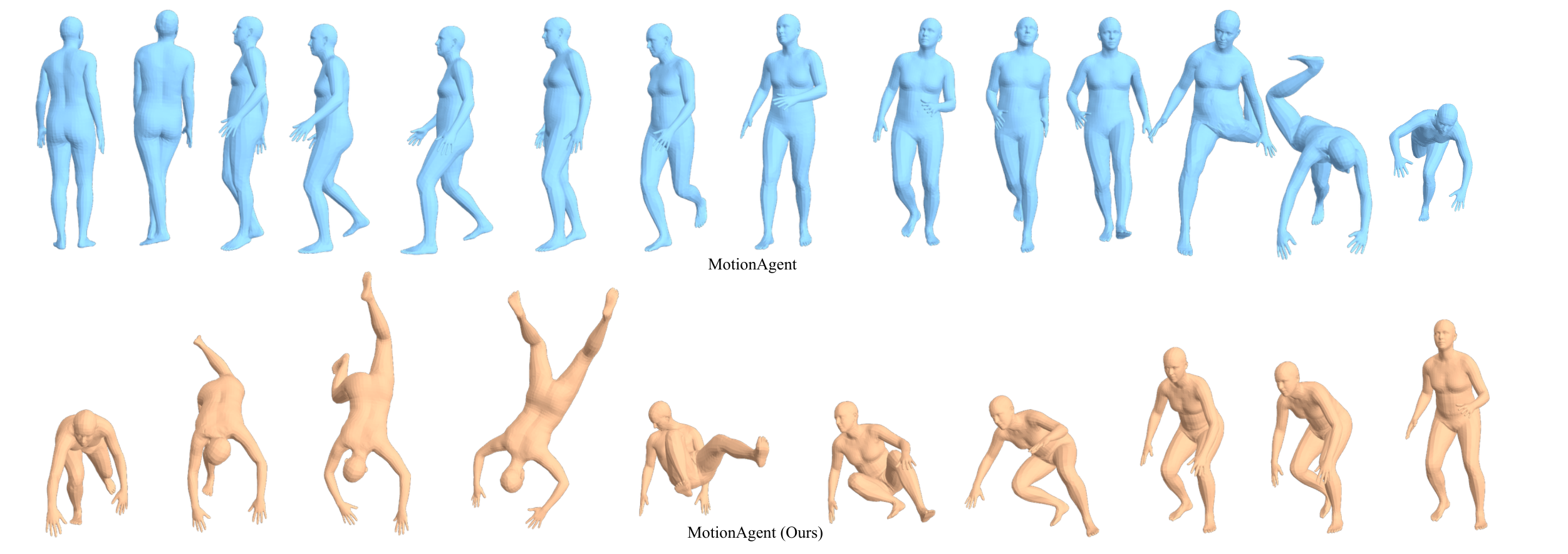}
\includegraphics[width=\linewidth]{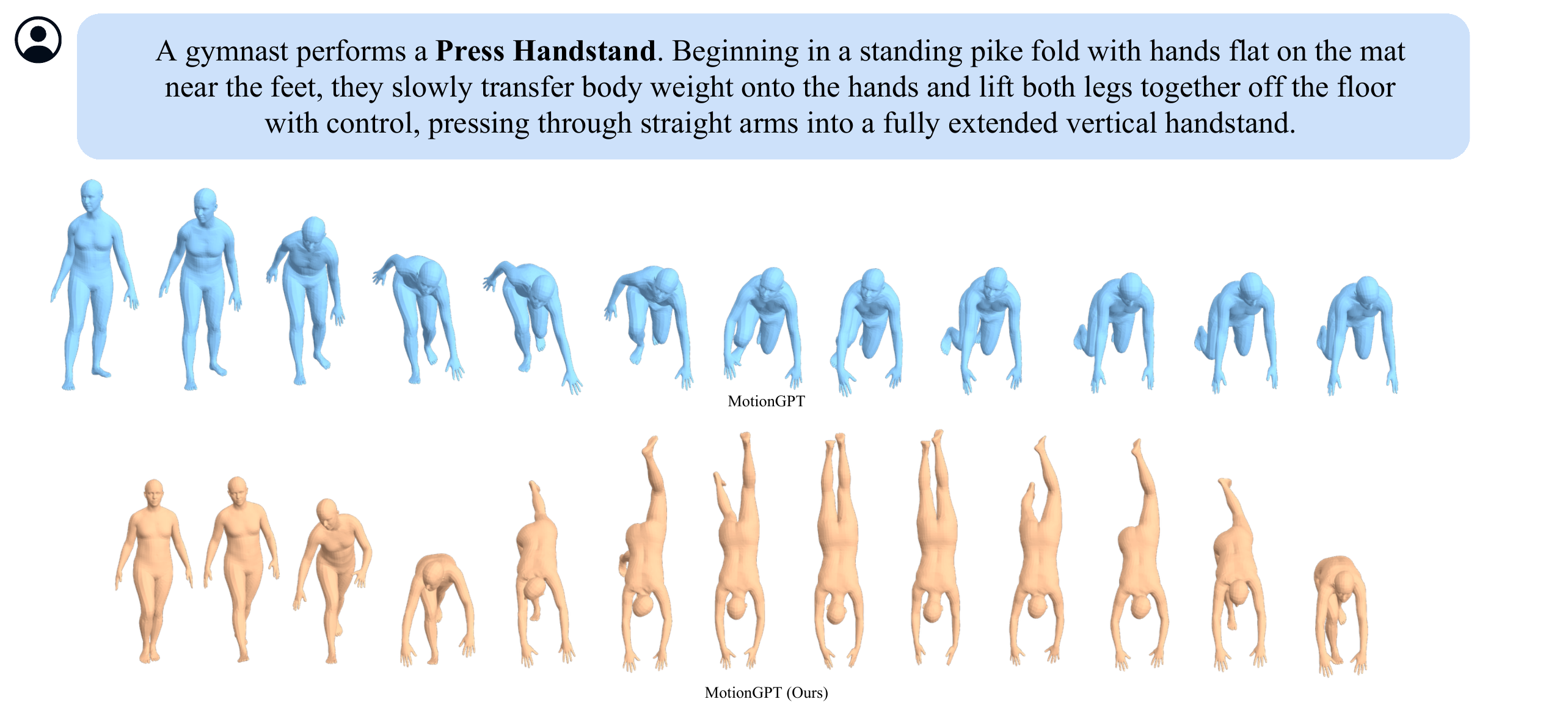}
\includegraphics[width=\linewidth]{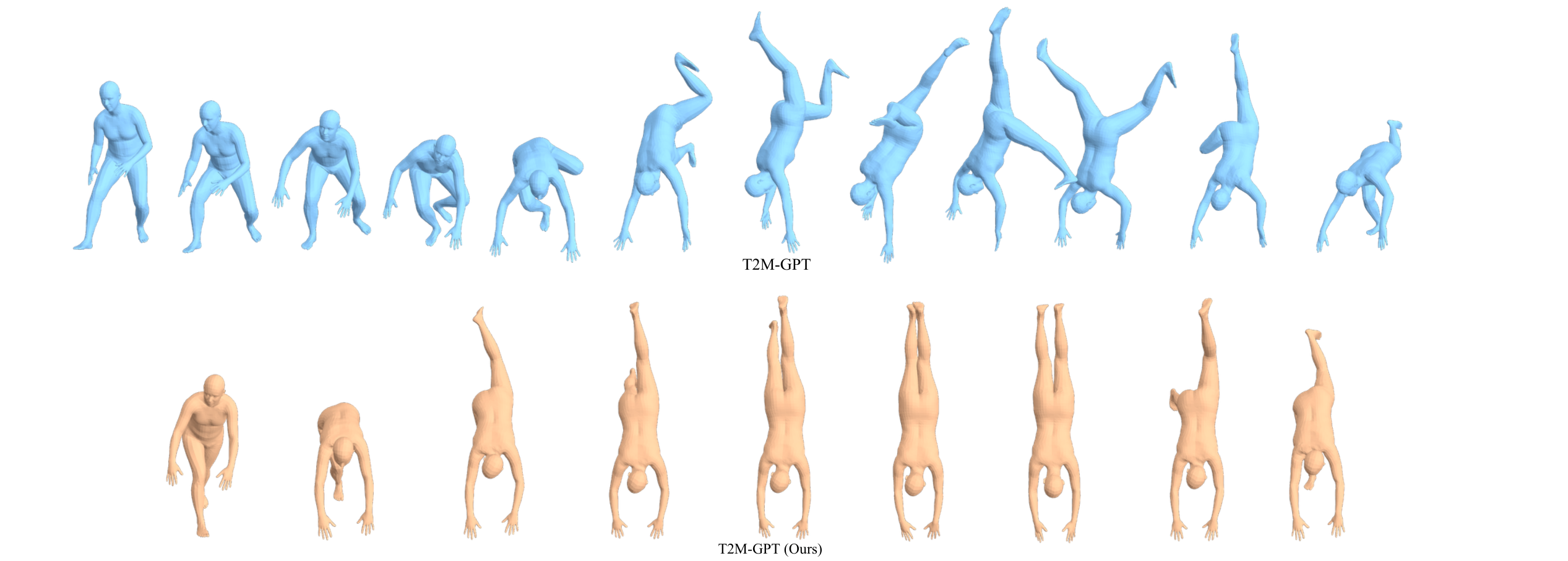}
\includegraphics[width=\linewidth]{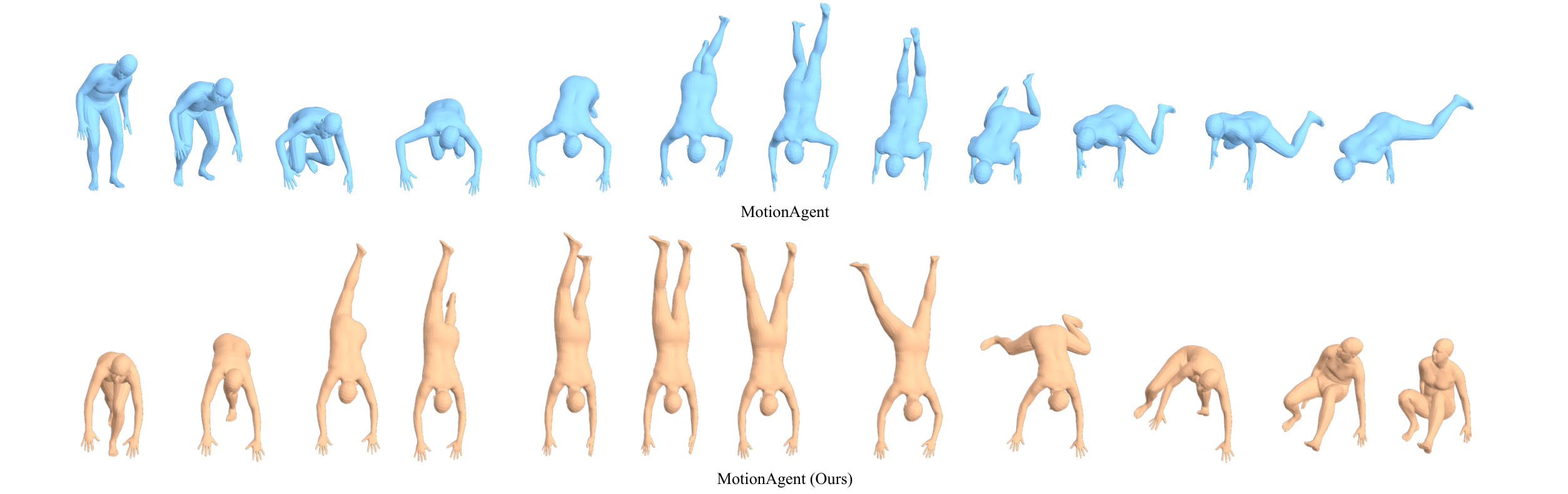}
\end{center}

\section{Per-Bone Coverage Statistics} 
\label{app:bone_coverage}

\begin{table}[h]
\centering
\small
\setlength{\tabcolsep}{6pt}

\begin{tabular}{cl ccc}
\toprule
\# & Bone & JS div. & L1 TV & Expansion \\
\midrule
1  & Pelvis $\to$ L\_Hip          & 1.03\% & 11.18\% & 12.94\% \\
2  & L\_Hip $\to$ L\_Knee         & 2.88\% & 17.48\% & 10.92\% \\
3  & L\_Knee $\to$ L\_Ankle       & 5.82\% & 27.00\% & 13.30\% \\
4  & L\_Ankle $\to$ L\_Foot       & 8.10\% & 30.81\% &  7.71\% \\
5  & Pelvis $\to$ R\_Hip          & 1.13\% & 11.79\% & 13.46\% \\
6  & R\_Hip $\to$ R\_Knee         & 2.68\% & 17.27\% & 10.12\% \\
7  & R\_Knee $\to$ R\_Ankle       & 5.55\% & 26.10\% & 12.95\% \\
8  & R\_Ankle $\to$ R\_Foot       & 7.86\% & 30.48\% &  6.06\% \\
9  & Pelvis $\to$ Spine1          & 1.24\% & 12.11\% & 15.05\% \\
10 & Spine1 $\to$ Spine2          & 3.34\% & 22.32\% &  3.87\% \\
11 & Spine2 $\to$ Spine3          & 2.35\% & 17.83\% &  3.09\% \\
12 & Spine3 $\to$ Neck            & 1.75\% & 16.20\% &  7.30\% \\
13 & Neck $\to$ Head              & 2.32\% & 18.04\% &  1.31\% \\
14 & Spine3 $\to$ L\_Collar       & 1.58\% & 14.46\% &  8.54\% \\
15 & L\_Collar $\to$ L\_Shoulder  & 1.75\% & 12.14\% &  1.24\% \\
16 & L\_Shoulder $\to$ L\_Elbow   & 2.55\% & 19.25\% &  1.91\% \\
\textbf{17} & \textbf{L\_Elbow $\to$ L\_Wrist}      & \textbf{3.81\%} & \textbf{22.44\%} & \textbf{0.14\%} \\
18 & Spine3 $\to$ R\_Collar       & 1.53\% & 13.90\% &  8.36\% \\
19 & R\_Collar $\to$ R\_Shoulder  & 1.45\% & 11.31\% &  0.67\% \\
20 & R\_Shoulder $\to$ R\_Elbow   & 1.77\% & 14.99\% &  1.98\% \\
21 & R\_Elbow $\to$ R\_Wrist      & 3.28\% & 21.20\% &  0.31\% \\
\midrule
   & \textbf{Mean}                & \textbf{3.04\%} & \textbf{18.49\%} & \textbf{6.72\%} \\
\bottomrule
\end{tabular}

\caption{\textbf{Per-bone distributional comparison across all 21 SMPL bones.}
JS divergence and L1 total variation measure the overall distributional
shift between HumanML3D and our synthetic data; expansion ratio measures
the fraction of $(\theta,\phi)$ region newly introduced by the synthetic
data, normalized by the original HumanML3D support.
All 21 bones exhibit non-zero expansion. The bone visualized in
Fig.~\ref{fig:distribution} (\textbf{L\_Elbow $\to$ L\_Wrist},
shown in bold) is highlighted and corresponds to one of the smallest expansions, indicating
that Fig.~\ref{fig:distribution} is a conservative example.}
\label{tab:bone_stats}
\end{table}

For each of the 21 bones in the SMPL skeleton, we compute the bone
vector (child joint $-$ parent joint) per frame, convert it to
spherical coordinates $(r, \theta, \phi)$, and discard $r$ to retain
only the direction $(\theta \in [-180^\circ, 180^\circ], \phi \in
[-90^\circ, 90^\circ])$.
We bin each bone's direction at $3^\circ \times 3^\circ$ resolution,
yielding a $121 \times 61$ spherical histogram that describes the
empirical distribution of directions the bone takes.
All statistics in Tab.~\ref{tab:bone_stats} are computed between two
such histograms: one aggregated over HumanML3D, one over our
synthetic data.

\paragraph{JS divergence and L1 total variation.}
JS divergence and L1 total variation jointly measure how far the
synthetic distribution has shifted from the original.
JS divergence is defined as
\begin{equation}
\mathrm{JS}(P \| Q) = \tfrac{1}{2}\mathrm{KL}(P \| M) + 
\tfrac{1}{2}\mathrm{KL}(Q \| M), 
\quad M = \tfrac{1}{2}(P+Q),
\end{equation}
where $P$ and $Q$ are the normalized direction histograms of the
synthetic and original data respectively.
JS is symmetric and bounded by $\ln 2 \approx 0.693$ under natural log,
making per-bone values directly comparable.
L1 total variation is defined as $\tfrac{1}{2} \sum_i |P_i - Q_i|$,
bounded by $1$.
The two are complementary: JS is sensitive to differences in
low-probability regions, while L1 TV measures the total probability
mass that must be moved to align the two distributions.
We report both so that the conclusion does not depend on a single
distance choice.

\paragraph{Expansion ratio.}
Expansion ratio measures coverage growth rather than distributional
shift.
Let $\mathcal{S}_{\text{orig}} = \{i : Q_i > 0\}$ denote the set of
$(\theta, \phi)$ cells covered by HumanML3D, and
$\mathcal{S}_{\text{aug only}} = \{i : P_i > 0 \wedge Q_i = 0\}$ denote
the cells reached only by the synthetic data.
We define
\begin{equation}
\mathrm{Expansion} \;=\; 
\frac{|\mathcal{S}_{\text{aug only}}|}
{\max(|\mathcal{S}_{\text{orig}}|, 1)}.
\end{equation}
Unlike JS divergence and L1 TV, which describe how the two
distributions differ within their joint support, the expansion ratio
directly answers whether the synthetic data enters new regions of
direction space that HumanML3D never reaches.
A non-zero value implies coverage growth rather than redistribution
within the existing support.

Tab.~\ref{tab:bone_stats} reports all three statistics for the 21
bones.
The \textbf{terminal joints} (wrist, ankle) consistently show the
largest JS and TV, while the \textbf{proximal joints} (pelvis-to-hip,
pelvis-to-spine) show the largest expansion ratios, reflecting that
synthetic motion both modifies extremity dynamics and opens up new
torso configurations not present in MoCap.


\section{Codebook Size Ablation}
\label{app:codebook_ablation}

\begin{table}[h]
\centering

\begin{tabular}{ccccccc}
\toprule
codebook & FID $\downarrow$ & Top-1 $\uparrow$ & Top-3 $\uparrow$ & Matching $\downarrow$ & Diversity & Rank \\
\midrule
1024  & $0.107 \pm 0.001$ & 0.490 & 0.777 & 3.106 & 9.666 & 3 \\
2048  & $0.094 \pm 0.001$ & 0.490 & 0.778 & 3.105 & 9.685 & 2 \\
4096  & $\mathbf{0.093 \pm 0.001}$ & 0.488 & 0.778 & 3.098 & 9.562 & 1 \\
8192  & $0.120 \pm 0.001$ & 0.488 & 0.778 & 3.116 & 9.600 & 5 \\
16384 & $0.126 \pm 0.001$ & 0.481 & 0.773 & 3.147 & 9.632 & 6 \\
32768 & $0.112 \pm 0.001$ & 0.458 & 0.745 & 3.273 & 9.467 & 4 \\
\bottomrule
\end{tabular}
\caption{\textbf{Codebook size ablation.} Lower FID and MM-Dist are better; higher Top-1 and Top-3 are better. 
The row with the lowest FID is shown in bold. 
All experiments are conducted with a HumanML3D-to-synthetic-data ratio of $1\mathord{:}1$.}
\vspace{2mm}
\vspace{-0.15in}
\label{tab:nb_code_ablation}
\end{table}
In this section, we ablate the codebook size $K$, which controls the
capacity of the discrete motion vocabulary.
The results are provided in Tab.~\ref{tab:nb_code_ablation}.
We adopt $K{=}2048$ based on three observations.
First, $K{=}2048$ achieves a reconstruction FID within $0.001$ of the
best $K{=}4096$ while strictly outperforming both $K{\le}1024$ and
$K{\ge}8192$, indicating it lies near the optimum of the FID curve.
Second, $K{=}2048$ keeps approximately $76\%$ of codes active, whereas
codebooks with $K{\ge}4096$ enter an under-utilized regime where most
codes are rarely accessed.
Third, $K{=}2048$ is the largest size at which both reconstruction FID
and codebook utilization continue to improve under our augmented data,
making it the operating point our synthesis pipeline can effectively
fill.

\section{Mixing Ratio Ablation}
\label{app:ratio_ablation}

\begin{table}[h]
\centering
\small
\begin{tabular}{lcc|cccc|c}
\toprule
& \multicolumn{2}{c|}{HumanML3D (in-dist)} & \multicolumn{4}{c|}{Motion-X++ MPJPE (OOD) $\downarrow$} & Codebook \\
Tokenizer & FID $\downarrow$ & Top-1 $\uparrow$ & Overall & Daily & Sports & Dance & util. \\
\midrule
T2M-GPT       & 0.132          & 0.499          & 272.5          & 251.2          & 314.4          & 183.8          & --- \\
\midrule
ratio=1:4       & 0.081 & 0.497          & 255.7          & 239.1          & \textbf{269.6} & 191.5          & \textbf{78.5\%} \\
\textbf{ratio=1:2(Ours)} & 0.076 & \textbf{0.504}       & \textbf{252.4} & \textbf{232.0} & 282.7          & \textbf{161.8} & 76.4\% \\
ratio=1:1        & \textbf{0.046} & \textbf{0.504} & 278.7          & 255.4          & 315.3          & 196.9          & 76.7\% \\
\bottomrule
\end{tabular}
\caption{Ablation on the mixing ratio between real HumanML3D data and synthetic data
($K{=}2048$, 500k iterations). Synthetic augmentation consistently improves in-
distribution reconstruction over the original T2M-GPT tokenizer, while different ratios
trade off reconstruction fidelity against out-of-distribution generalization on Motion-
X++. The ratio $1{:}2$ gives the best overall OOD performance and is used in our final
model.}
\label{tab:hml_ablation}
\end{table}

Tab.~\ref{tab:hml_ablation} highlights the trade-off between fidelity and coverage
when mixing real and synthetic training data.
Original MoCap data, HumanML3D, is indispensable for preserving accurate in-distribution reconstruction, as it provides precise MoCap supervision and
anchors the tokenizer to realistic motion structure. At the same time, synthetic data also significantly improves in-distribution reconstruction quality over the original tokenizer baseline, showing that its benefit is not limited to out-of-distribution
generalization. This suggests that the synthetic set does not merely add diversity, but also helps the tokenizer better cover underrepresented yet still valid motion patterns within the HumanML3D distribution. 

Synthetic data, in contrast, primarily contributes
coverage: as its proportion increases, out-of-distribution performance on Motion-X++ improves, indicating that the tokenizer benefits from exposure to a broader motion support than HumanML3D alone can offer. 

These results support our central design
choice: the most effective tokenizer is obtained not by replacing real data with
synthetic data, but by balancing scarce high-quality MoCap data with large-scale
synthetic augmentation.

\section{Data Generation Implementation Details}
\label{app:impl} 
\label{app:data_gen_impl}

\subsection{Hierarchical Representation and Limb Replacement}
Our evolutionary pipeline operates on single HumanML3D poses in the
$22{\times}3$ joint representation.
The skeleton is modeled as a rooted kinematic tree with pelvis as the
root, followed by two leg chains, one spine chain, and two arm chains.
Rather than exchanging an arbitrary set of joints in Cartesian space,
we perform crossover on a hierarchical representation of
\emph{non-torso} bone directions.
Concretely, we extract 13 exchangeable limb bones
(head, collar--shoulder, shoulder--elbow, elbow--wrist,
hip--knee, knee--ankle, and ankle/foot bones) and represent each bone
in a local coordinate system defined by the torso and its parent limb.

To perform limb replacement, we first choose a \emph{subtree root}
uniformly from a fixed set of joints whose descendants correspond to a
replaceable limb segment:
pelvis, left/right hip, left/right knee, upper spine, left/right
collar, left/right shoulder, and left/right elbow.
Each selected root is mapped to a predefined subtree of limb bones.
For example, choosing the pelvis replaces both lower-body chains,
choosing a collar replaces the corresponding arm branch, and choosing
an elbow replaces only the forearm and wrist segment.
We then swap the local bone directions in that subtree between the two
parent poses.
This hierarchical replacement preserves the kinematic structure of the
motion and avoids implausible Cartesian mixing across unrelated body
parts.

\subsection{Crossover and Mutation}
Before applying crossover, each pose is converted from the stored joint
order to the coordinate order expected by the angle-limit code and then
mapped into the local bone representation.
Given a pair of parents, crossover exchanges the selected subtree of
local bone directions between them, producing two candidate children.
Each child is reconstructed on the other parent's limb lengths, so that
the swapped branch keeps a coherent morphology after replacement.

Mutation is applied \emph{only} to the bones affected by crossover.
For each swapped bone, with probability $\texttt{MRL}$ we sample a
random rotation axis from $\{x,y,z\}$ and a rotation angle from a
zero-mean Gaussian with standard deviation $\texttt{SDL}$ degrees, and
rotate the local bone vector accordingly.
This differs from the main-text description in two ways:
first, mutation is not applied to the whole body but only to the
replaced subtree; second, the perturbation is defined in the local
coordinate system, so it changes relative articulation instead of
global orientation.
After reconstruction, we optionally apply a global rigid rotation, but
this branch is disabled in our final pipeline.
Finally, each accepted pose is translated vertically so that its lowest
joint lies on a sampled ground plane.

\subsection{Pose Prior and Validity Filtering}
We use the pose prior of Akhter and Black~\cite{akhter2015pose} as a
filter on synthesized poses.
In the original prior, anatomically valid bone directions are defined
through pose-conditioned joint-angle limits.
For first-level limb bones directly attached to the torso, the prior
stores occupancy masks over discretized spherical angles.
For second-level bones, validity is conditioned on the orientation of
the parent limb through a separating plane and a bounded region in a
local 2D coordinate system.
Following the released implementation, the spherical angles are
discretized at $3^\circ$ resolution, with $\theta\in[-180^\circ,
180^\circ]$ and $\phi\in[-90^\circ, 90^\circ]$.

To use this prior on HumanML3D, we build three torso-centered local
frames: one for the upper body, one for the lower body, and one for the
collar extension introduced by the 22-joint skeleton.
Each non-torso bone is then expressed in the appropriate local frame and
checked against the original angular constraints.
The resulting validity score is simply the number of valid bones among
the 13 exchangeable ones.
In our filter, a synthesized child is discarded only if its score is
both below a threshold $\texttt{Th}$ and lower than its parent's score;
equivalently, we keep a child whenever it is at least as valid as its
parent, or already exceeds the threshold.
In the final pipeline we use $\texttt{Th}{=}13$, under which
$67.33\%$ of generated poses are retained.

\subsection{Evolution Hyperparameters}
Our released code exposes the full evolutionary process through a small
set of hyperparameters.
The number of generations is controlled by $\texttt{G}$, parent
selection by $\texttt{F}$, local mutation by $\texttt{MRL}$ and
$\texttt{SDL}$, global mutation by $\texttt{MRG}$ and
$\texttt{SDG}$, and validity filtering by $\texttt{Th}$.
When $\texttt{Mer}{=}\texttt{True}$, newly synthesized poses are merged
back into the current population instead of replacing it.

In the configuration used for our main data-generation pipeline, we set
$\texttt{G}{=}11$, $\texttt{F}{=}0.2$, $\texttt{MRL}{=}0.8$,
$\texttt{SDL}{=}50^\circ$, $\texttt{MG}{=}\texttt{False}$,
$\texttt{Th}{=}13$, and $\texttt{seed}{=}111$.
We run the pipeline in the \texttt{largefiles} mode, where data are
grouped across source files and evolved independently.
With $\texttt{F}{=}0.2$, each generation selects $20\%$ of the current
population as fathers and another $20\%$ as mothers, yielding up to
$0.4N$ candidate children before filtering for a population of size
$N$.
Because $\texttt{Mer}{=}\texttt{True}$, these survivors are appended to
the current population, so the dataset grows over generations instead
of being refreshed from scratch.

\subsection{Interpolation}
After pose synthesis, we densify the generated set by interpolation in
joint-rotation space rather than by linear interpolation on joint
coordinates.
For each input file, we split the poses into two halves and pair them
in order; if the number of poses is odd, we pad the smaller half with
the first pose so that all pairs are matched.
Each pose is converted to per-joint quaternions by inverse kinematics,
and intermediate poses are generated by spherical linear interpolation
(SLERP) over joint rotations.
In our implementation, the root translation is kept fixed to the first
pose in the pair, while bone lengths are inherited from that first pose
during forward kinematics reconstruction.
We use $T{=}65$ interpolated frames for each pair.

\subsection{Training Details}
For motion tokenization, we set a temporal downsampling rate $l = 4$, hidden dimension 512, code dimension 512 and three residual blocks. Due to the introduction of large-scale synthetic motion data, we increase the codebook size from the original $K = 512$ to $K = 2048$ to accommodate the expanded motion diversity. To balance the original HumanML3D data and the synthetic data during training, we apply the weighted sampling strategy with an effective mixing ratio of 1:1. 
We evaluate the effectiveness of the augmented tokenizer on downstream motion generation models, including T2M-GPT~\cite{zhang2023t2mgpt}, MotionGPT~\cite{jiang2023motiongpt}, and MotionAgent~\cite{wu2025motionagent}. For all models, we replace the original motion tokenizer with our augmented tokenizer while keeping the rest of the architecture unchanged. During training, motion sequences are first encoded into discrete tokens using the learned codebook then we follow the original training protocols of each baseline model for fair comparison. Both tokenizer training and downstream model finetuning are conducted on a single NVIDIA RTX 6000 Pro GPU.

\section{Evaluation Metrics and Motion Representation Details}
\label{app:eval}

\paragraph{Mean Per Joint Position Error (MPJPE).}
MPJPE measures the reconstruction accuracy of the tokenizer at the joint level.
Given a motion sequence of $T$ frames with $J$ joints, let $\mathbf{p}_{t,j} \in \mathbb{R}^3$ and $\hat{\mathbf{p}}_{t,j} \in \mathbb{R}^3$ denote the ground-truth and reconstructed 3D position of joint $j$ at frame $t$, respectively.
MPJPE is defined as:
\begin{equation}
    \text{MPJPE} = \frac{1}{T \cdot J} \sum_{t=1}^{T} \sum_{j=1}^{J} \left\| \mathbf{p}_{t,j} - \hat{\mathbf{p}}_{t,j} \right\|_2
\end{equation}
A lower MPJPE indicates more accurate reconstruction. We report MPJPE in millimeters (mm).

\paragraph{Fréchet Inception Distance (FID).}
FID measures the distributional similarity between generated motions and real motions.
Let $\mu_{\text{gt}}$, $\Sigma_{\text{gt}}$ and $\mu_{\text{pred}}$, $\Sigma_{\text{pred}}$ denote the mean and covariance of $f_{\text{gt}}$ and $f_{\text{pred}}$, respectively.
FID is computed as:
\begin{equation}
    \text{FID} = \left\| \mu_{\text{gt}} - \mu_{\text{pred}} \right\|^2
    + \mathrm{Tr}\!\left( \Sigma_{\text{gt}} + \Sigma_{\text{pred}} - 2\left(\Sigma_{\text{gt}} \Sigma_{\text{pred}}\right)^{\frac{1}{2}} \right)
\end{equation}
where $\mathrm{Tr}(\cdot)$ denotes the matrix trace. A lower FID indicates that the generated motion distribution is closer to the real distribution.

\paragraph{R-Precision.}
R-Precision evaluates the consistency between generated motions and their text descriptions via a retrieval task.
Given a generated motion, we rank its distance to a batch of 32 text descriptions (1 matching + 31 mismatched) using the feature-level Euclidean distance.
R-Precision at Top-$k$ ($k \in \{1, 2, 3\}$) reports the fraction of cases where the correct text description appears in the Top-$k$ retrieved results.
Higher R-Precision indicates better text-motion alignment.

\paragraph{Multimodal Distance (MM-Dist).}
MM-Dist measures the average feature-level distance between generated motions and their corresponding text descriptions.
Given $N$ text-motion pairs, MM-Dist is defined as:
\begin{equation}
    \text{MM-Dist} = \frac{1}{N} \sum_{i=1}^{N} \left\| f_{\text{pred},i} - f_{\text{text},i} \right\|_2
\end{equation}
where $f_{\text{pred},i}$ and $f_{\text{text},i}$ are the motion and text features of the $i$-th pair.
A lower MM-Dist indicates better semantic alignment between text and motion.

\paragraph{Diversity.}
Diversity measures the variance of generated motions across the dataset.
We randomly sample $S_{\text{dis}}$ pairs of generated motions, and compute:
\begin{equation}
    \text{Diversity} = \frac{1}{S_{\text{dis}}} \sum_{i=1}^{S_{\text{dis}}} \left\| f_{\text{pred},i} - f'_{\text{pred},i} \right\|_2
\end{equation}
where $f_{\text{pred},i}$ and $f'_{\text{pred},i}$ are the features of the $i$-th sampled pair.
Following~\cite{guo2022generating}, we set $S_{\text{dis}} = 300$.
A higher Diversity indicates that the model generates a wider variety of motions.

\section{Per-Subset Reconstruction on Motion-X++}
\label{app:per_subset}

Tab.~\ref{tab:tok-ood-full} reports per-frame MPJPE (mm) on all eight
Motion-X++ subsets, comparing T2M-GPT, our tokenizer trained with a 1:1 real-to-synthetic ratio, and our final tokenizer trained with a 1:2 real-to-synthetic ratio.
Our final tokenizer improves over T2M-GPT on \textbf{7 of 8 subsets},
with the only regression on $\textit{perform}$ ($+2.8\%$, 923 samples,
$3.6\%$ of total).
We attribute this to $\textit{perform}$ being relatively well-covered
by HumanML3D, leaving little room for synthetic data to contribute.
By contrast, the $\text{ratio}{=}1:1$ variant — which over-fits to
in-distribution FID — improves on only 4 of 8 subsets and degrades
overall MPJPE relative to T2M-GPT, confirming the trade-off discussed
in Sec.~\ref{sec:exp_tokenizer}.

\begin{table}[h]
\centering
\small
\setlength{\tabcolsep}{6pt}

\begin{tabular}{lr c cc}
\toprule
& & \multicolumn{3}{c}{MPJPE (mm) $\downarrow$} \\
\cmidrule(lr){3-5}
Group & $n$ & T2M-GPT & $\text{ratio=1:1}$ & \textbf{Ours} ($\text{ratio}{=}1:2$) \\
\midrule
animation & 559     & 352.7 & 337.7\,{\scriptsize\textcolor{ForestGreen}{($-4.3\%$)}} & 344.1\,{\scriptsize\textcolor{ForestGreen}{($-2.5\%$)}} \\
haa500    & 6{,}944  & 322.7 & 315.3\,{\scriptsize\textcolor{ForestGreen}{($-2.3\%$)}} & \textbf{282.7}\,{\scriptsize\textcolor{ForestGreen}{($-14.1\%$)}} \\
humman    & 971     & 372.7 & 382.3\,{\scriptsize\textcolor{red}{($+2.6\%$)}}        & \textbf{358.3}\,{\scriptsize\textcolor{ForestGreen}{($-4.0\%$)}} \\
idea400   & 12{,}042 & 244.7 & 255.4\,{\scriptsize\textcolor{red}{($+4.4\%$)}}        & \textbf{232.0}\,{\scriptsize\textcolor{ForestGreen}{($-5.5\%$)}} \\
kungfu    & 1{,}032  & 438.4 & 443.2\,{\scriptsize\textcolor{red}{($+1.1\%$)}}        & \textbf{419.0}\,{\scriptsize\textcolor{ForestGreen}{($-4.6\%$)}} \\
music     & 3{,}394  & 180.8 & 196.9\,{\scriptsize\textcolor{red}{($+8.9\%$)}}        & \textbf{161.8}\,{\scriptsize\textcolor{ForestGreen}{($-11.7\%$)}} \\
perform   & 923     & 262.3 & 280.5\,{\scriptsize\textcolor{red}{($+7.0\%$)}}        & 269.7\,{\scriptsize\textcolor{red}{($+2.7\%$)}} \\
\midrule
\textbf{Overall} & \textbf{25{,}865} & \textbf{272.7} & 278.7\,{\scriptsize\textcolor{red}{($+2.2\%$)}} & \textbf{252.4}\,{\scriptsize\textcolor{ForestGreen}{($-8.1\%$)}} \\
\bottomrule
\end{tabular}
\caption{\textbf{Per-frame MPJPE (mm) on all Motion-X++ subsets.}
The two rightmost columns report MPJPE alongside the relative change
$\Delta$ over T2M-GPT
(\textcolor{ForestGreen}{green}: improvement,
\textcolor{red}{red}: regression).
$\text{ratio}{=}1:1$ is the variant with best in-distribution FID;
$\text{ratio}{=}1:2$ is our final tokenizer.
Lower MPJPE is better.}
\label{tab:tok-ood-full}
\end{table}


\end{document}